\documentclass[sigconf]{acmart}

\AtBeginDocument{%
  \providecommand\BibTeX{{%
    \normalfont B\kern-0.5em{\scshape i\kern-0.25em b}\kern-0.8em\TeX}}}

\copyrightyear{2022}
\acmYear{2022}
\setcopyright{acmcopyright}\acmConference[KDD '22]{Proceedings of the 28th ACM SIGKDD Conference on Knowledge Discovery and Data Mining}{August 14--18, 2022}{Washington, DC, USA}
\acmBooktitle{Proceedings of the 28th ACM SIGKDD Conference on Knowledge Discovery and Data Mining (KDD '22), August 14--18, 2022, Washington, DC, USA}
\acmPrice{15.00}
\acmDOI{10.1145/3534678.3539481}
\acmISBN{978-1-4503-9385-0/22/08}
\usepackage{subfig}

\usepackage{microtype}
\usepackage{float}
\usepackage{array}
\usepackage{booktabs}
\usepackage{multirow}
\usepackage{subfig}
\usepackage{graphicx}
\usepackage{booktabs} % for professional tables
\usepackage{hyperref}

\usepackage{amsmath}

\usepackage{amssymb}
\usepackage{mathtools}
\usepackage{amsthm}
\usepackage{algorithm}
\usepackage{algorithmicx}
\usepackage{algpseudocode}
\usepackage{bbm}
\usepackage{balance}
\usepackage[capitalize,noabbrev]{cleveref}

\usepackage[textsize=tiny]{todonotes}

\begin{document}

\title{S2RL: Do We Really Need to Perceive All States in Deep Multi-Agent Reinforcement Learning?}

\author{Shuang Luo}
\email{luoshuang@zju.edu.cn}
\authornote{Both authors contributed equally to this research. This work was completed while Shuang Luo was a member of the Huawei Noah’s Ark Lab for Advanced Study.}
\affiliation{%
  \institution{Zhejiang University}
  \city{}
  \state{Hangzhou}
  \country{China}
}

\author{Yinchuan Li}
\email{liyinchuan@huawei.com}
\authornotemark[1]
\affiliation{%
  \institution{Huawei Noah’s Ark Lab}
  \city{}
  \state{Beijing}
  \country{China}
}

\author{Jiahui Li}
\email{jiahuil@zju.edu.cn}
\affiliation{%
  \institution{Zhejiang University}
  \city{}
  \state{Hangzhou}
  \country{China}
}

\author{Kun Kuang}
\email{kunkuang@zju.edu.cn}
\authornote{Kun Kuang and Chao Wu are the corresponding authors.}
\affiliation{%
  \institution{Zhejiang University}
  \city{}
  \state{Hangzhou}
  \country{China}
}

\author{Furui Liu}
\email{liufurui2@huawei.com}
\author{Yunfeng Shao}
\email{shaoyunfeng@huawei.com}
\affiliation{%
  \institution{Huawei Noah’s Ark Lab}
  \city{}
  \state{Beijing}
  \country{China}
}

\author{Chao Wu}
\email{chao.wu@zju.edu.cn}
\authornotemark[2]
\affiliation{%
  \institution{Zhejiang University}
  \city{}
  \state{Hangzhou}
  \country{China}
}

\renewcommand{\shortauthors}{Shuang Luo et al.}
\newcommand{\etal}{\textit{et al}.}
\newcommand{\ie}{\textit{i}.\textit{e}.}
\newcommand{\eg}{\textit{e}.\textit{g}.}
\begin{abstract}
Collaborative multi-agent reinforcement learning (MARL) has been widely used in many practical applications, where each agent makes a decision based on its own observation. 
Most mainstream methods treat each local observation as an entirety when modeling the decentralized local utility functions. However, they ignore the fact that local observation information can be further divided into several entities, and only part of the entities is helpful to model inference. Moreover, the importance of different entities may change over time. 
To improve the performance of decentralized policies, the attention mechanism is used to capture features of local information. Nevertheless, existing attention models rely on dense fully connected graphs and cannot better perceive important states. To this end, we propose a \emph{sparse state based MARL} (S2RL) framework, which utilizes a sparse attention mechanism to discard irrelevant information in local observations.
The local utility functions are estimated through the self-attention and sparse attention mechanisms separately, then are combined into a standard joint value function and auxiliary joint value function in the central critic.
We design the S2RL framework as a plug-and-play module, making it general enough to be applied to various methods. Extensive experiments on StarCraft II show that S2RL can significantly improve the performance of many state-of-the-art methods.

% \blfootnote{$^*$ Both authors contributed equally to this research.}
% \blfootnote{$^\dagger$ Kun Kuang and Chao Wu are the corresponding authors.}

\end{abstract}





\maketitle

\section{Introduction}

Multi-agent reinforcement learning (MARL) provides a framework for multiple agents to solve complex sequential decision-making problems, with broad applications including robotics control~\cite{li2021shapley,robot}, video gaming~\cite{vinyals2019grandmaster,liu2019}, traffic light control~\cite{trafic2017} and autonomous driving~\cite{kiran2021deep,autonomous}. In the paradigm of centralized training with decentralized execution (CTDE)~\cite{MADDPG,QMIX}, each local agent models a policy that treats the local observation as input. 
% In deep reinforcement learning, the local policies are modeled by a neural network usually implemented by recurrent neural networks.
% However, most mainstream methods take local observations as input while underestimate the role of entities.
However, the role of entities is underestimated by most mainstream methods.
Entities are defined as fine-grained tokens of observations, \eg, \emph{obstacles, landmarks, enemies}, which determine the inference process of the model. Specifically, they treat all entities observed as a whole and contribute indiscriminately to the estimation of the value function. But in some cases, the importance of each entity changes dynamically over time steps.

% However, most of the mainstream methods take the local observations as inputs while underestimate the role of entities. Entities are defined as the fine-grained tokens of the observation, \eg  \emph{enemy $1$ location, agent type, health, etc} in StarCraft II, which determine the inference procedures of models. Specifically, they treat all entities of an observation as a whole and indiscriminate their contributions to the estimation of value functions. While in some scenarios, the importance of the each entity changes dynamically along with the time step.
\begin{figure*}[!t]
    \centering
    \subfloat[Six friendly Hydralisks face 8 enemy Zealots.]{\includegraphics[scale=0.50]{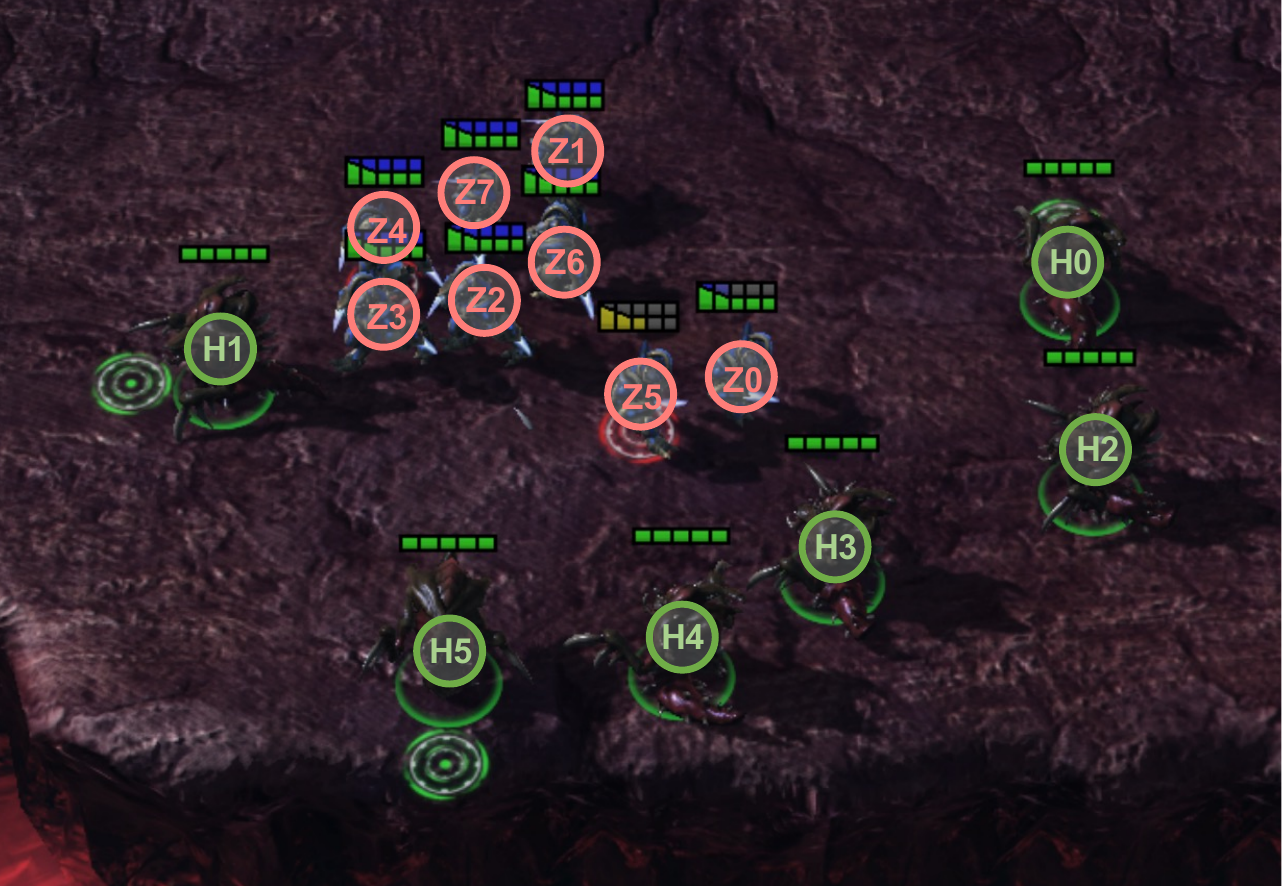}\label{fig:ex1}}\;    
    \subfloat[Dense attention distribution]{\includegraphics[scale=0.305]{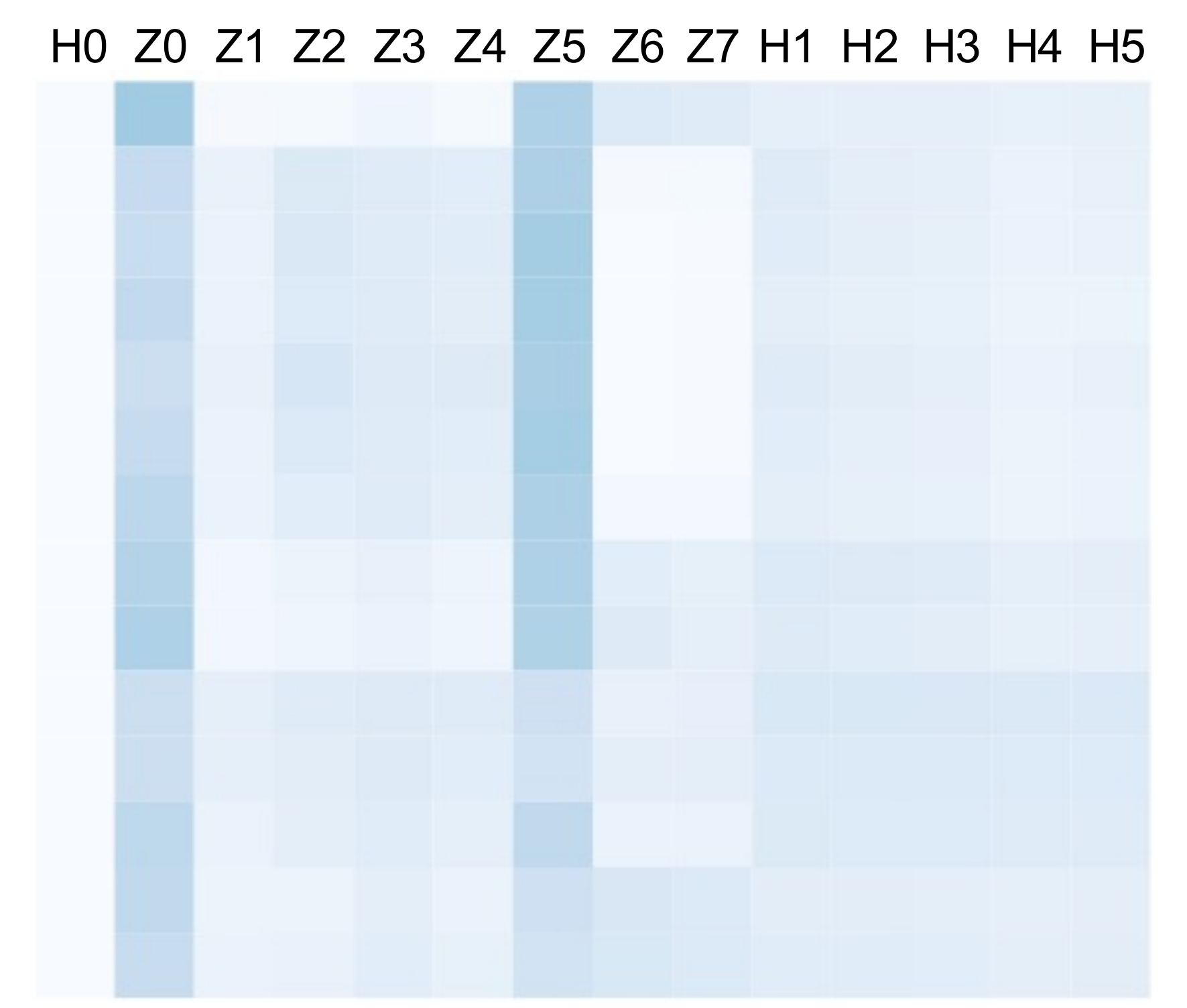}\label{fig:ex2}}
    \subfloat[Sparse attention distribution]{\includegraphics[scale=0.305]{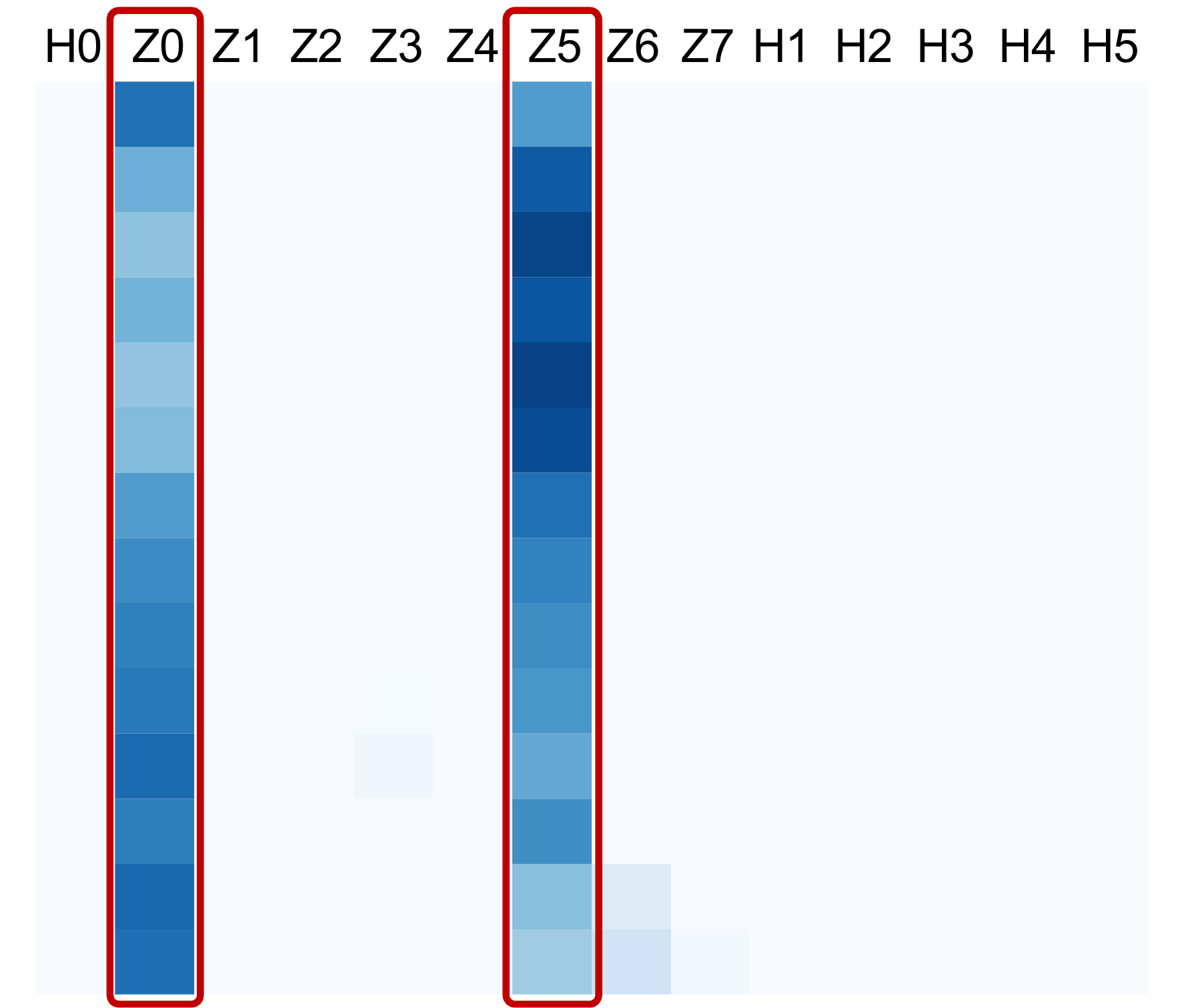}\label{fig:ex3}}
    \caption{A visualization example of agent performance on the StarCraft II super-hard scenario $6h\_vs\_8z$. As shown in (a), the closest to the green agent H3 are the red enemies Z0 and Z5. Thus the corresponding policy is that H3 only needs to focus on Z0 and Z5, which are more likely to be annihilated. (b) shows the softmax attention distribution of the H3 observations, finding that some weights are still assigned to irrelevant entities. In contrast, the sparse attention in (c) only focuses on Z0 and Z5.}
    \label{fig:ex}
\end{figure*}

To better leverage the observation information, the attention mechanism has been adopted~\cite{attention} for its ability to learn the interaction relationship among entities and dynamically focus on the crucial parts. Most existing attention mechanisms compute importance weights based on dense fully-connected graphs, where all participants are assigned scores according to their contribution to model decisions. In practice, however, not all entities are helpful for model inference, and discarding redundant entities can sometimes improve overall performance. Therefore, it is crucial for agents to learn to select valuable observations and exclude others. 
To better illustrate this phenomenon, a visualization of the StarCraft II scene and the corresponding attention distribution is shown in Figure~\ref{fig:ex}. The green agent H3 is very close to the red enemies Z0 and Z5. Hence agent H3 focusing only on enemies Z0 and Z5 is more effective. However, from the traditional dense attention distribution of H3, we can see that H3 assigns much attention to irrelevant entities. Note that the large state space brings great difficulties to policy learning for a more complex MARL environment.

An ideal way to solve this issue is to replace the traditional attention mechanism with sparse attention. 
% \textcolor{red}{For example, SparseMAAC~\cite{sparsemaac} extends MAAC~\cite{MAAC} with sparsity by replacing the softmax activation function in attention mechanism with $\gamma$-sparsemax directly. 
% It calculates the importance of different agents observations and actions, while we mainly focus on selecting key entities of individual observations.}
From Figure~\ref{fig:ex}(c), we can see that adopting sparse attention can well guide H3 only to perceive Z0 and Z5, reducing the observation space that the agent needs to perceive.  
However, simply applying sparse attention to local agents will corrupt the training. The main reason is that the network cannot distinguish which entity is more important at the beginning of training. If the agents only focus on critical entities initially, it may lead to an inadequate exploration of the environment and thus converge to a suboptimal policy. More specifically, temporarily discarding some entities can be seen as a policy exploration behavior. Meanwhile, local policies need to execute their exploration strategies. When these two strategies are executed simultaneously, it is difficult for the model to converge.

% , and discard some entities tentatively can be deemed as a kind of exploration strategy.
% An ideal way to solve this issue is to replace the traditional attention mechanism with sparse attention. However, simply apply sparse attention on the local agents will corrupt the training. The main reason is that when the training starts, the network can not distinguish which entity is more important, and discard some entities tentatively can be deemed as a kind of exploration strategy.
% At the same time, the local policy need to execute its exploration strategy. A model can hardly convergences when these two strategy are performed simultaneously.

In this paper, we propose a \emph{Sparse State based MARL}~(S2RL) framework, where the sparse attention mechanism is utilized as the auxiliary for guiding the local agent training.
% Specifically, we first model the local value function using a traditional self-attention mechanism. Then, we construct a corresponding utility function for each agent, which is implemented by a sparse attention mechanism.
% Finally, in the central critic, the local value function and utility function respectively form the joint value function and auxiliary value function, which are further used to train the entire network.
In particular, we model the local utility function using a traditional self-attention mechanism. Then, we construct a corresponding auxiliary utility function for each agent, which is implemented by a sparse attention mechanism. The local utility and auxiliary utility functions respectively form the joint value and auxiliary value functions, which are further used to train the entire network.
Since the sparse attention mechanism is considered auxiliary and thus does not corrupt the training process,  the auxiliary value function is also used to update the entire framework. To this end, local agents can learn patterns to focus on essential entities while ignoring redundancy.

% In this paper, we propose a \emph{Sparse State based MARL} (S2RL) framework where the spare attention mechanism is utilized as the auxiliary for guiding the training of local agents.
% Concretely, we first model a local value function with traditional  self-attention mechanism. 
% Meanwhile, we construct a corresponding utility function for each agent which is implemented with spare attention mechanism.
% Finally, in the central critic, and the local value functions and utility functions form the joint value function and auxiliary value function respectively which are further used to train the whole network.
% The sparse attention mechanism is treated as an auxiliary hence will not corrupt the learning procedure. Since the auxiliary value function is also used to updated the whole framework, the local agents can learn the pattern to focus on important entities while neglecting the redundant.
% Agent observation space gradually increases with the size of the population, causing great difficulty for policy learning. Sparse attention mechanism reduces the number of entities that agent needs to pay attention to, simplifying the exploration space of the intelligent body to some extent.

% The implementation details will be introduced in \emph{Section}~\ref{sec4}.

% \subsection{Main Contributions}

Our main contributions are summarized as follows:
\begin{itemize}
\item To the best of our knowledge, this paper is the first attempt that uses enhanced awareness of crucial states as the auxiliary in MARL to improve convergence rate and performance. 
% And reinforcement learning cannot only use crucial states, which prevents the algorithm from converging.
\item We propose the S2RL framework for local agents to perceive crucial states while preserving all states. The proposed framework thus addresses the inability to converge using only a small number of partial observations.
\item We design the S2RL framework as a plug-and-play module, making it general enough to be applied to various methods in the CTDE paradigm.
\item The extensive experiments on StarCraft II show that S2RL brings remarkable improvements to existing methods, especially in complicated scenarios.
\end{itemize}

The remainder of the paper is organized as follows.
In Section 2, we introduce the background of MARL and the CTDE framework. In Section 3, we propose our S2RL framework. Experimental results are presented in Section 4. Related works are presented in Section 5. Section 6 concludes the paper.

\begin{figure*}[!t]
    \centering
    \includegraphics[scale=1]{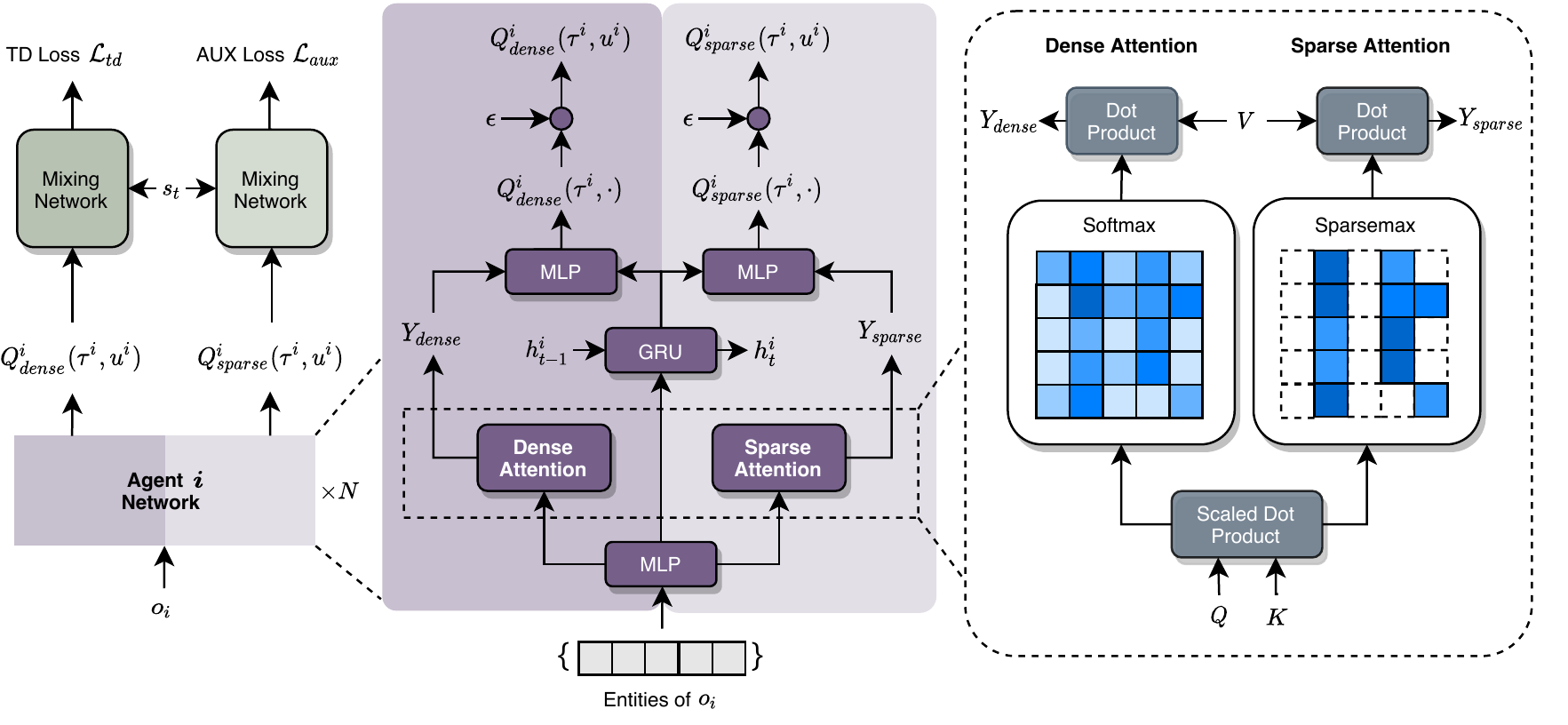}
    \caption{Illustration of the proposed S2RL method. We use the value-based MARL framework under the CTDE paradigm and apply the S2RL method to an agent utility network. The core of S2RL is composed of the dense attention module and sparse attention module, where sparse attention serves as an auxiliary for guiding the dense attention training.}
    \label{fig:method}
\end{figure*}

% Illustration of the proposed \emph{Sparse State based MARL} (S2RL) method. We use the value-based MARL framework under the CTDE paradigm and apply the S2RL module in agent utility network. The core of S2RL is utilizing sparse attention module as auxiliary for guiding the training of dense attention. The weight distribution maps of softmax is dense while the distribution maps of sparsemax is more compact.

\section{PRELIMINARIES}
\subsection{Dec-POMDP}
A fully cooperative multi-agent sequential task can be described as a decentralized partially observable Markov decision process (Dec-POMDP)~\cite{Dec-POMDP}, which is canonically formulated by the tuple:
\begin{align}
M=<\mathcal{I}, \mathcal{S}, \mathcal{U}, P, R, \Omega, G, \gamma >.
\end{align}
In the process, $\mathcal{I}\equiv\{1,2,\dots,N\}$ is the finite set of agents and $s \in \mathcal{S}$ represents the global state of the environment.
At each time step, each agent $i \in \mathcal{I}$ receives an individual partial observation $o^{i} \in {\Omega}$ according to the observation function $G(s,i)$ and selects an action $u^i \in \mathcal{U}$, forming a joint action $\boldsymbol{u}$. 
This results in a transition to the next state $s'$ according to the state transition function $P(s'|s, \boldsymbol{u}): \mathcal{S} \times \mathcal{U} \times \mathcal{S} \rightarrow \left[0,1\right]$. 
All agents share the same global reward $r$ based on the reward function $ R(s, \boldsymbol{u}): \mathcal{S} \times \mathcal{U} \rightarrow \mathbb{R}$, and $\gamma \in \left[0,1\right)$ is the discount factor. Due to partially observable setting, each agent has an action-observation history $\tau^{i} \in \mathcal{T} \equiv(\Omega \times \mathcal{U})^{*}$ and learns its individual policy $\pi^{i}\left(u^i | \tau^{i}\right)$ to jointly maximize the discounted return. The joint policy $\boldsymbol{\pi}$ induces a joint action-value function: $Q^{tot}_{\boldsymbol{\pi}}(s, \boldsymbol{u})=\mathbb{E}_{s_{0: \infty}, \boldsymbol{u}_{0: \infty}}\left[\sum_{t=0}^{\infty} \gamma^{t} r_{t} \mid s_{0}=s, \boldsymbol{u}_{0}=\boldsymbol{u}, \boldsymbol{\pi}\right]$.

\subsection{CTDE Framework}
The centralized training and decentralized execution (CTDE) is a popular paradigm used in deep multi-agent reinforcement learning~\cite{RIAL,QMIX,NCC,QPD,COMA}, which enables agents to learn their individual policies in a centralized way. During the centralized training process, the learning model can access the state and provide global information to assist the agents in exploring and training. However, each agent only makes decisions based on its local action-observation history during decentralized execution.
% Centralized training with decentralized execution(CTDE) means that agents learn in a centralized training way via communicating among others or with access to additional global state information. But during the decentralized execution phase, each agent makes its decision conditioned only on local action-observation history.
The Individual-Global-Max principle~\cite{QTRAN} guarantees the consistency between joint and local greedy action selections. Agents can obtain the optimal global reward by maximizing the individual utility function of each agent. Thus a more robust individual value function can benefit the whole team in cooperative multi-agent tasks. 

The global Q-function $Q^{tot}_{\boldsymbol{\pi}}$ is calculated by all individual value functions: $Q^{to t}_{\boldsymbol{\pi}}(\boldsymbol{\tau}, \mathbf{u})=F([Q_{\pi}^{i}(\tau^{i}, u^{i})]_{i=1}^N,s;\theta)$
% $F\left(\left[Q_{\pi^i}^{i}\left(\tau^{i}, u^{i}\right)\right]_{i}^{N}\right)$
, where $\boldsymbol{\tau} \equiv \mathcal{T}^{n}$ is a joint action-observation history and $\boldsymbol{u}$ is a joint action, $F$ is the credit assignment function parameterized by $\theta$ to learn value function factorization. Each agent learns its own utility function by maximizing the global value function $Q^{tot}$, which is trained end-to-end to minimise the following TD loss:
\begin{align}
    \mathcal{L}(\theta) =  \mathbb{E}_{\mathcal{D}} \left[\left(y^{tot} - Q^{tot}(\boldsymbol{\tau},\boldsymbol{u},s;\theta)\right)^2\right],
\end{align}
where $\mathcal{D}$ is the replay buffer, $y^{tot}=r+\gamma\max_{\boldsymbol{u}'}Q^{tot}(\boldsymbol{\tau}',\boldsymbol{u}',s';\theta^-)$ and $\theta^-$ is the parameter of the target network~\cite{DQN}.

\section{Sparse state based MARL}\label{sec4}

In this section, we propose a novel sparse state based MARL framework that is general enough to be plugged into any existing value-based multi-agent algorithm.
% In the follows, we detail the proposed S2RL  method as shown in Figure~\ref{fig:method}. 
As shown in Figure~\ref{fig:method}, our framework adopts the CTDE paradigm, where each agent learns its individual utility network by optimizing the TD loss of the mixing network. During the execution, the mixing network is removed, and each agent acts according to its local policy derived from its value function. 
Distinguish from other value-based methods, our agents’ value functions or policies carry out the process of selection and discrimination according to the importance of different entities of state.
To enable efficient and effective learning among agents between different entities of state, our method be described by three steps: 
% \textcolor{red}
{1) selection; 2) discrimination; 3) learning.}

\subsection{Selection and Discrimination}
It is a dynamic process to assign attentions based on the contribution of the observed entities to the value estimation.
In our framework, we adopt the self-attention module ~\cite{attention} to capture the relational weights between the observed entities of the agents.
% Assigning different attention to entities of observation according to their contributions for  estimation of value functions is a dynamic process. In our framework, we adopt the self-attention module~\cite{attention} to capture the relationship weights between agent' observation entities.
In particular, an agent $i$ observes $M$ other entities at time step $t$, then the corresponding input of utility network $O_t^i$ is defined as $O_t^i = [{\boldsymbol{o}^{i, 1}_t},\ldots,{\boldsymbol{o}^{i, M}_t}]^T \in \mathbb{R}^{{M} \times d_E}$ with $d_E$ being the entity dimension and $o^{i, m}_t \in \mathbb{R}^{d_E}$ being the state information of the $m$-th ($m\in \{1,...,M \}$) entity. 
All observed entities are embedded to $d_X$ dimension via an embedding function $f(\cdot): \mathbb{R}^{d_E} \rightarrow \mathbb{R}^{d_X}$ as follows:
\begin{align}
X_{t}^{i}=\left\{f(\boldsymbol{o}_{t}^{i, 1}), \ldots, f(\boldsymbol{o}_{t}^{i, M})\right\},~i \in \mathcal{I}.
\end{align}
% All observed entities are embedded to $d_X$ dimension via an embedding function $f(\cdot): \mathbb{R}^{d_E} \rightarrow \mathbb{R}^{d_X}$, denoted as
% $X_{t}^{i}=\left\{f(\boldsymbol{o}_{t}^{i, 1}), \ldots, f(\boldsymbol{o}_{t}^{i, M})\right\},~i \in \mathcal{I}.$
% % follows:
% % \begin{align}
% % X_{t}^{i}=\left\{f(\boldsymbol{o}_{t}^{i, 1}), \ldots, f(\boldsymbol{o}_{t}^{i, M})\right\},~i \in \mathcal{I}.
% % \end{align}
Then, the embedding feature of each agent ${X} \in \mathbb{R}^{{M} \times d_X}$ is projected to query $Q = X {W}_{Q}$, key $K = X {W}_{K}$ and value $V = X {W}_{V}$ representation, where $\{{W}_Q, {W}_K, {W}_V\} \in \mathbb{R}^{d_X \times d_X}$ are trainable weight matrices.
Then, $Q$, $K$, $V$ are input into the self-attention layer to calculate the entities importance for the model decision, which is given by
\begin{align}
 \operatorname{Attn}(Q, K, V)= \operatorname{softmax}\left(\frac{Q K^{T}}{\sqrt{d_{X}}}\right) V.
\end{align}

One of the limitations of the softmax activation function is that the resulting probability weights for any element never appear to be zero, which further leads to dense output probabilities. 
% However, in reinforcement learning, it is crucial for agents to learn to select valuable observations, which reduces the number of entities that agent needs to pay attention to, simplifying the exploration space.
Nevertheless, for the sake of simplifying the exploration space and selecting valuable observations, it is crucial for agents to reduce the number of entities to focus on.
Hence, a sparse probability distribution is desired to distinguish between critical and irrelevant entities, which can	accelerate convergence and improve performance.
To start with, inspired by sparsemax~\cite{sparsemax,entmax}, we consider introducing sparse states to enhance the perception of valuable entities of agent observation and neglect the others.

% A limitation of the softmax activation function is that the resulting probability weights of any element never appear to be zero, resulting in dense output probabilities. In practice, it is crucial for agents to learn to select the valuable observation information, which implies a sparse probability distribution is desired to discriminate between actual states and secondary. Thus we introduce sparsemax~\cite{sparsemax} as an alternative to replace the softmax in the attention mechanism. 
% Sparsemax commonly defines a threshold below which small probability values are truncated to zero. 
We denote the products of the query with all keys ${Q K^{T}}$ as $Z \in \mathbb{R}^{M \times M}$, which consists of $M$ rows $\{{\boldsymbol{z}_1}, \ldots, {\boldsymbol{z}_M}\}$ with ${\boldsymbol{z}_m} \in \mathbb{R}^{M}$ being the logits of the $m$-th row.
Afterwards, we define a matrix sorting operator $\operatorname{SortMat}(\cdot)$ as follows:
\begin{align}
\widetilde{Z} = \operatorname{SortMat}(Z) = [\operatorname{SortVec}(\boldsymbol{z}_1)^T,\dots,\operatorname{SortVec}(\boldsymbol{z}_M)^T]^T,
\end{align}
where $\operatorname{SortVec}(\cdot)$ sorts the elements of vector in descending order. 
% For example, $\operatorname{SortVec}(\boldsymbol{z}_m) = \{z\in [\boldsymbol{z}_m] \mid z_{j} > z_{k},~\text{if}~j > k\}$.
Then we calculate
\begin{align}
\boldsymbol{n}(\widetilde{Z}) = [n_1,\dots,n_M]^T,
\end{align}
where $n_m := \max{\{n \in [M] \mid 1+n \widetilde{Z}_{m,n} > \sum_{k \leq n} \widetilde{Z}_{m,k} \}}$ is the maximal number of crucial elements in $\boldsymbol{z}_m$ that we intend to preserve, while other elements is set to zero in the subsequent operations. 
We define
\begin{align}
\boldsymbol{c} = [c_{1},\cdots,c_{M}]^T
\end{align}
with $c_{m} = \sum_{k \leq n_m} Z_{m,k} -1$ and the scaling vector as
\begin{align}
\boldsymbol{\mu} = [\frac{1}{n_1},\cdots,\frac{1}{n_M}]^T.
% \begin{bmatrix} 
% \frac{1}{n_1} & 0 & \cdots & 0 \\ 
% 0 & \frac{1}{n_2} & \cdots  & 0 \\
% \vdots & \vdots & \ddots  & \vdots \\
% 0 & 0 & \cdots  & \frac{1}{n_M}
% \end{bmatrix}.
\end{align}
Then, the threshold matrix is calculated as
\begin{align}
\Delta = \boldsymbol{1}_{M\times 1} (\boldsymbol{c} \odot \boldsymbol{\mu})^T,
\end{align}
where $\boldsymbol{1}_{M\times 1} \in \mathbb{R}^{M}$ is an all-one vector and $\odot$ is the pointwise product. The sparse attention weights matrix $P$ is obtained by
\begin{align}
P = [Z - \Delta]_{+},
\end{align}
where $[\cdot]_{+}:= \max \{0, \cdot \}$.
% The element in $Z$ above the corresponding coordinate in $\Delta$ will be shifted by this amount, and the others less will be truncated to zero.
Thus, the sparse attention is given by
\begin{align}
 \operatorname{sAttn}(Q, K, V)= P V,
\end{align}
which can retain most of the essential properties of softmax while assigning zero probability to low-scoring choices. Therefore, the model will pay more attention to critical entities when making decisions, reducing the attention to other redundant entities.

% Therefore, the model decision can based on the critical entities and other redundant entities are discarded.

% Sparsemax is a new activation function similar to the traditional softmax, but the difference is that it can output sparse probabilities. Thus we use sparsemax as activation function to replace the softmax in the attention mechanism. 

% $P=\operatorname{sparsemax}\left(\frac{Q K^{T}}{\sqrt{d_{k}}}\right) V$

% This sparse attention mechanism enables us to focus on essential state of token, which helps agents learn more efficiently and effectively. 

\begin{algorithm}[!tb]
\caption{Sparse State based MARL Algorithm}

\textbf{Initialize}: Critic network $\theta_{\rho}$, target critic $\hat{\theta}_{\rho}$ = $\theta_{\rho}$, agents' Q-value networks $\theta_{\pi} = (\theta_1, \dots, \theta_N)$ and Replay buffer $\mathcal{D}$
\begin{algorithmic}[1] 
\For{each training episode $e$}
\State $t = 0$, $s_0$ = initial state, ${o}_{0}^i=G({s}_{0}, i)$ and $h^i_0 = 0$ for $i \in \mathcal{N}$
%\STATE 
\While{$s \neq terminal$ and $t < T$} 
% \emph{For deep pMeta-RL only}
\State $t = t + 1$
\For{ each agent $i$}
\State Calculate dense attention $Y_{dense}$ by \eqref{Osta}
\State Calculate sparse attention $Y_{sparse}$ by \eqref{Oaux}
\State Calculate trajectory encode $h_t^i$ by \eqref{h}
\State Obtain $Q_{dense}^{i}(\tau^{i}, \cdot)$ by \eqref{Qsta}
\State Obtain $Q_{sparse}^{i}(\tau^{i}, \cdot)$ by \eqref{Qaux}
% \State $Q_{dense}^{i}\left(\tau^{i}, \cdot\right) = {agent}^i\left(Y_{dense}, h_t^i\right)$
% \State $Q_{sparse}^{i}\left(\tau^{i}, \cdot\right) = {agent}^i\left(Y_{sparse}, h_t^i\right)$
\State Sample $u^i$ from $\pi^i(Q_{dense}^i, \epsilon)$
\EndFor
\State Execute actions $\boldsymbol{u}_t = (u^1, \dots, u^N)$
\State Receive reward $r_{t+1}$ and next state ${s}_{t+1}$
\EndWhile
\State Store episodes in replay buffer $\mathcal{D}$
\State Sample a random minibatch of episodes from $\mathcal{D}$
\State \textbf{Dense Attention Loss:}
\State \quad \;\; Compute $\mathcal{L}_{td}(\theta_{\pi}, \theta_{\rho})$ by \eqref{Lsta}
\State \textbf{Auxiliary Sparse Attention Loss:}
\State \quad \;\; Compute $\mathcal{L}_{aux}(\theta_{\pi}, \theta_{\rho})$ by \eqref{Laux}
\State Update $\theta_{\pi}$ and $\theta_{\rho}$ by \eqref{totalloss}
\State Every $C$ episodes reset $\hat{\theta}_{\rho}$ = $\theta_{\rho}$
\EndFor
\end{algorithmic}
\label{algo}
\end{algorithm}

\subsection{Learning with Sparse Loss}
% The direct may to achieve sparse attention mechanism is to replace the traditional self-attention activation function with sparsemax. However, the model can not distinguish which entity is more important at the beginning of training. To mitigate this issue, we utilize sparse attention mechanism shown in the right side of Figure~\ref{fig:method} as auxiliary to guide  the training of local agents.  
Obviously, the sparse attention mechanism can be realized by directly replacing the traditional self-attention activation function with a sparse distribution function. However, the model cannot distinguish which entity is vital from the beginning. Thus directly adopting the sparse attention mechanism will have performance regression. To address this issue, we design the structure shown on the right side of Figure ~\ref{fig:method} to guide the training of local agents, where we utilize two routes to exploit dense and sparse attention, respectively. Dense attention guarantees that the algorithm can converge, while sparse attention is a powerful auxiliary to enhance the agent perception of critical entities, thereby improving performance.

To do this, the sparse attention module and dense attention module share the weight matrices $\{{W}_Q, {W}_K, {W}_V\}$ and the GRU module. Denote the parameters of these two networks by $\theta_{\pi}$.
The projected matrix $Q$ and $K$ are fed into both dense and sparse attention. Then, we calculate the weighted sum of $V$ to obtain the output 
\begin{equation}\label{Osta}
Y_{dense} = \operatorname{Attn}(Q, K, V) \in \mathbb{R}^{M \times d_X},
\end{equation}
\begin{equation}\label{Oaux}
Y_{sparse} = \operatorname{sAttn}(Q,K,V)  \in \mathbb{R}^{M \times d_X}.
\end{equation}
In our implementation, the GRU~\cite{GRU} module is utilized to encode an agent's history of observations and actions via
\begin{equation}\label{h}
h_t^i = GRU(X_t^i, h_{t-1}^i).
\end{equation}
%\nonumber
Then, $Y_{dense}$ and $Y_{sparse}$ are concatenated with the output of GRU separately to estimate the individual value function as follows:
\begin{equation}\label{Qsta}
Q_{dense}^{i}(\tau^{i}, \cdot) = \operatorname{Agent}^i(Y_{dense}, h_{t}^i),
\end{equation}
\begin{equation}\label{Qaux}
Q_{sparse}^{i}(\tau^{i}, \cdot) = \operatorname{Agent}^i(Y_{sparse}, h_{t}^i).
\end{equation}
% Action selection is performed by each agent maximizing $Q_{dense}^{i}$ and $Q_{sparse}^{i}$ for subsequent computation of centralized training. Moreover, the action selected by $Q_{dense}^{i}$ is executed in the  environment.
% For the exploration policy, $\epsilon$-greedy is adopted and the exploration rate of episode $\epsilon$ decreases over time.
% Action selection is performed by each agent to maximize $Q_{dense}^{i}$ and $Q_{sparse}^{i}$ for subsequent computations in centralized training. 
Each agent selects the action that maximizes $Q_{dense}^{i}$ and $Q_{sparse}^{i}$ for subsequent computations in centralized training. 
In addition, the action selected by $Q_{dense}^{i}$ is executed in the environment. For the exploration strategy, $\epsilon$-greedy is adopted, and the exploration rate of $\epsilon$ decreases over time.

To better learn the role of entities in credit assignment, we use a mixing network to estimate the global Q-values $Q_{dense}^{tot}$ and $Q_{sparse}^{tot}$, using per-agent utility $Q_{dense}^{i}$ and $Q_{sparse}^{i}$. Since the auxiliary estimation is calculated in the individual utility function, our proposed S2RL is seamlessly integrated with various valued-based algorithms. 
For example, we can use the mixing network, a feed-forward neural network introduced by QMIX~\cite{QMIX}. The mixing network mixes the agent network outputs monotonically.
The parameters of the mixing network parameterized by $\theta_{\rho}$ are conditioned on the global states and are generated by a hyper-network. Then, we minimize the following TD loss to update the dense attention module:
\begin{equation}\label{Lsta}
\mathcal{L}_{td}\left(\theta_{\pi}, \theta_{\rho}\right)=\mathbb{E}_{\mathcal{D}}\left[\left(r+\gamma \max _{\boldsymbol{u}^{\prime}} \bar{Q}_{dense}^{tot}\left(s^{\prime}, \boldsymbol{u}^{\prime}\right)-Q_{dense}^{tot}(s, \boldsymbol{u})\right)^{2}\right],
\end{equation}
where $\bar{Q}_{dense}^{tot}$ is the target network, and the expectation is estimated with uniform samples from the same replay buffer ${\mathcal{D}}$.
In the meanwhile, the AUX Loss is given by
\begin{equation}\label{Laux}
\mathcal{L}_{aux}\left(\theta_{\pi}, \theta_{\rho}\right)=\mathbb{E}_{\mathcal{D}}\left[\left(r+\gamma \max _{\boldsymbol{u}^{\prime}} \bar{Q}_{sparse}^{tot}\left(s^{\prime}, \boldsymbol{u}^{\prime}\right)-Q_{sparse}^{tot}(s, \boldsymbol{u})\right)^{2}\right],
\end{equation}
where $\bar{Q}_{tot}^{aux}$ is the auxiliary target network.
% and the expectation is estimated with uniform samples from the same replay buffer ${\mathcal{D}}$ as the one for the TD Loss.

\begin{figure*}[h]
    \centering
    \subfloat[3s5z]{\includegraphics[scale=0.26]{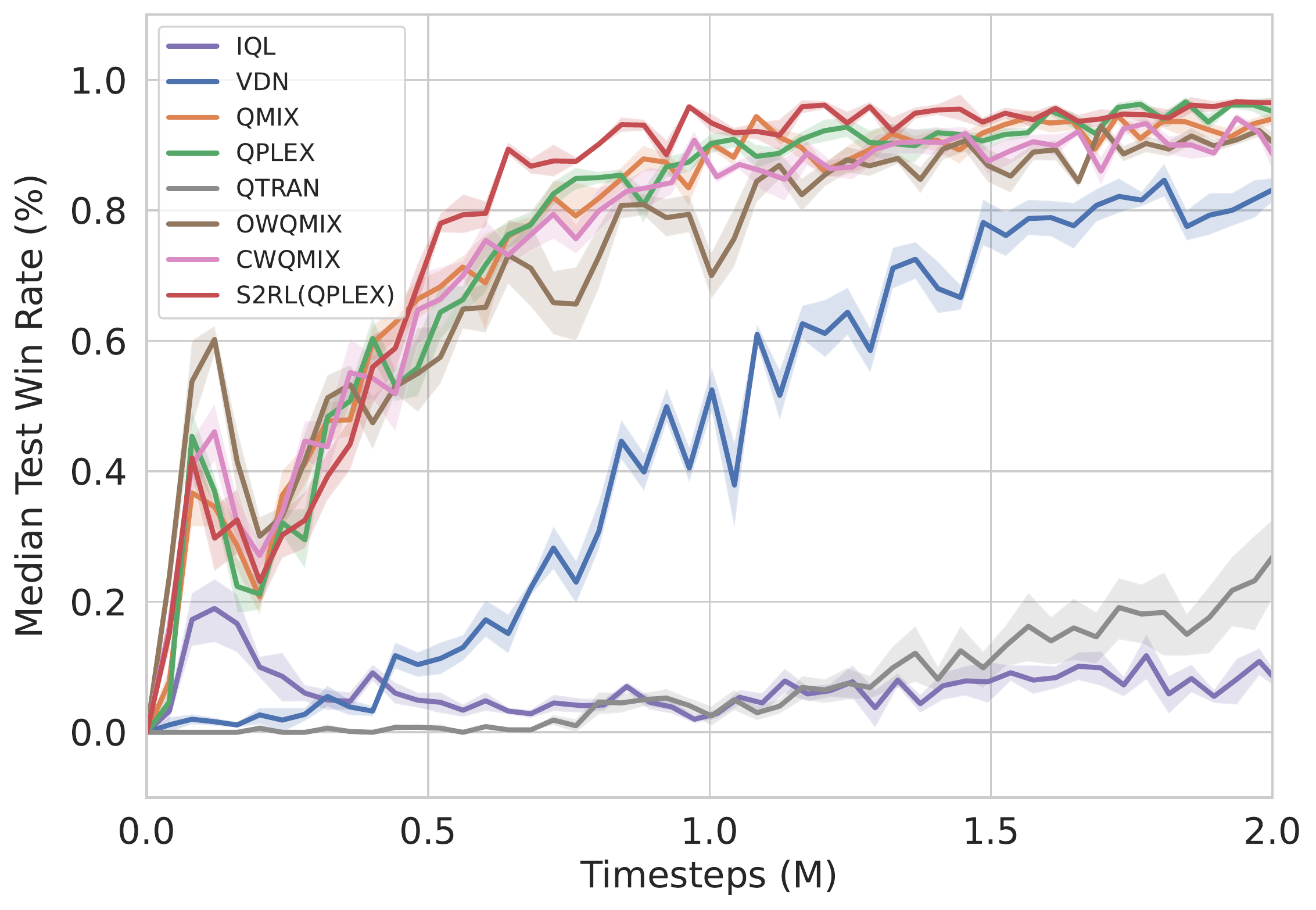}}
    \subfloat[3s\_vs\_5z]{\includegraphics[scale=0.26]{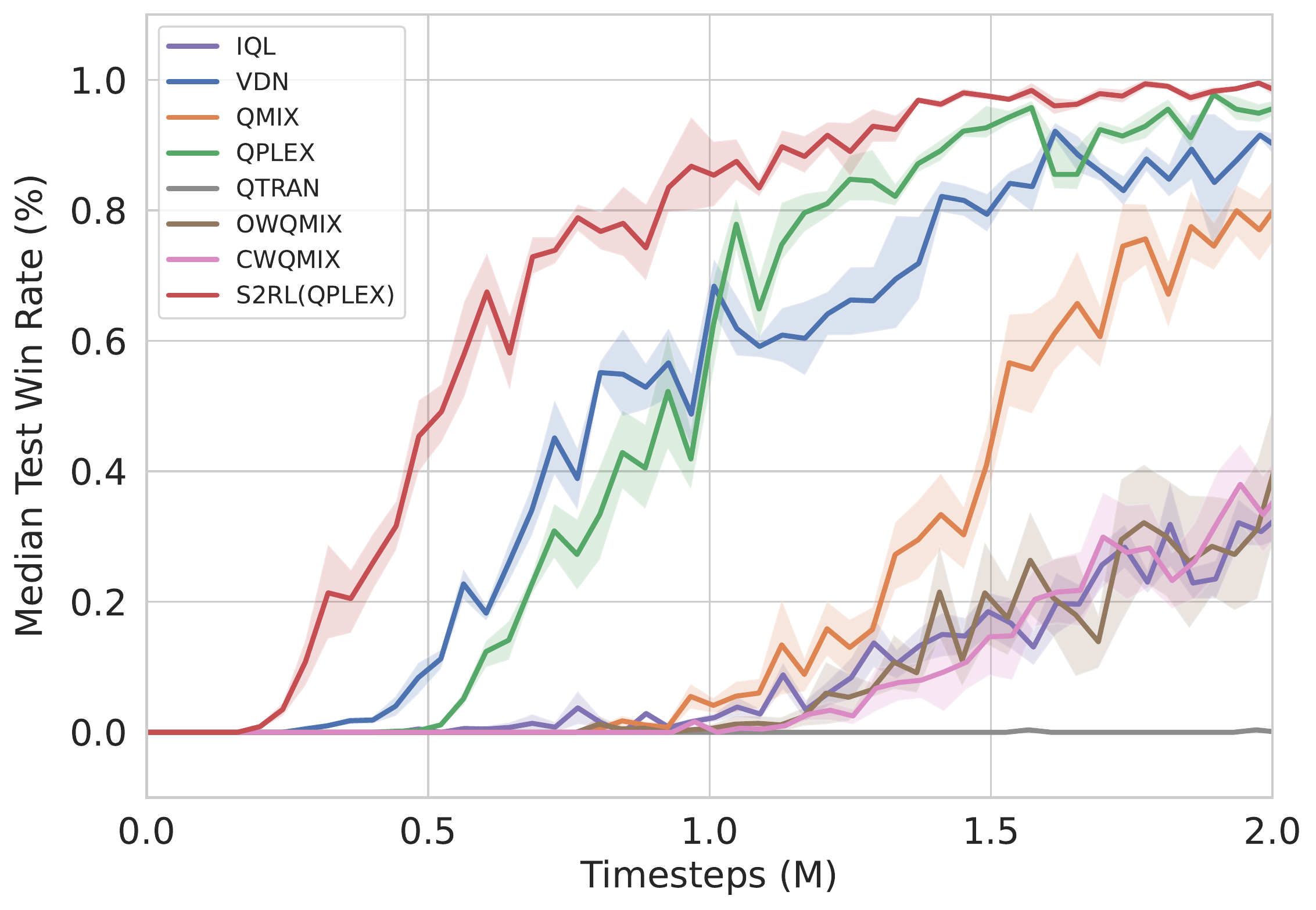}}
    \subfloat[5s10z]{\includegraphics[scale=0.26]{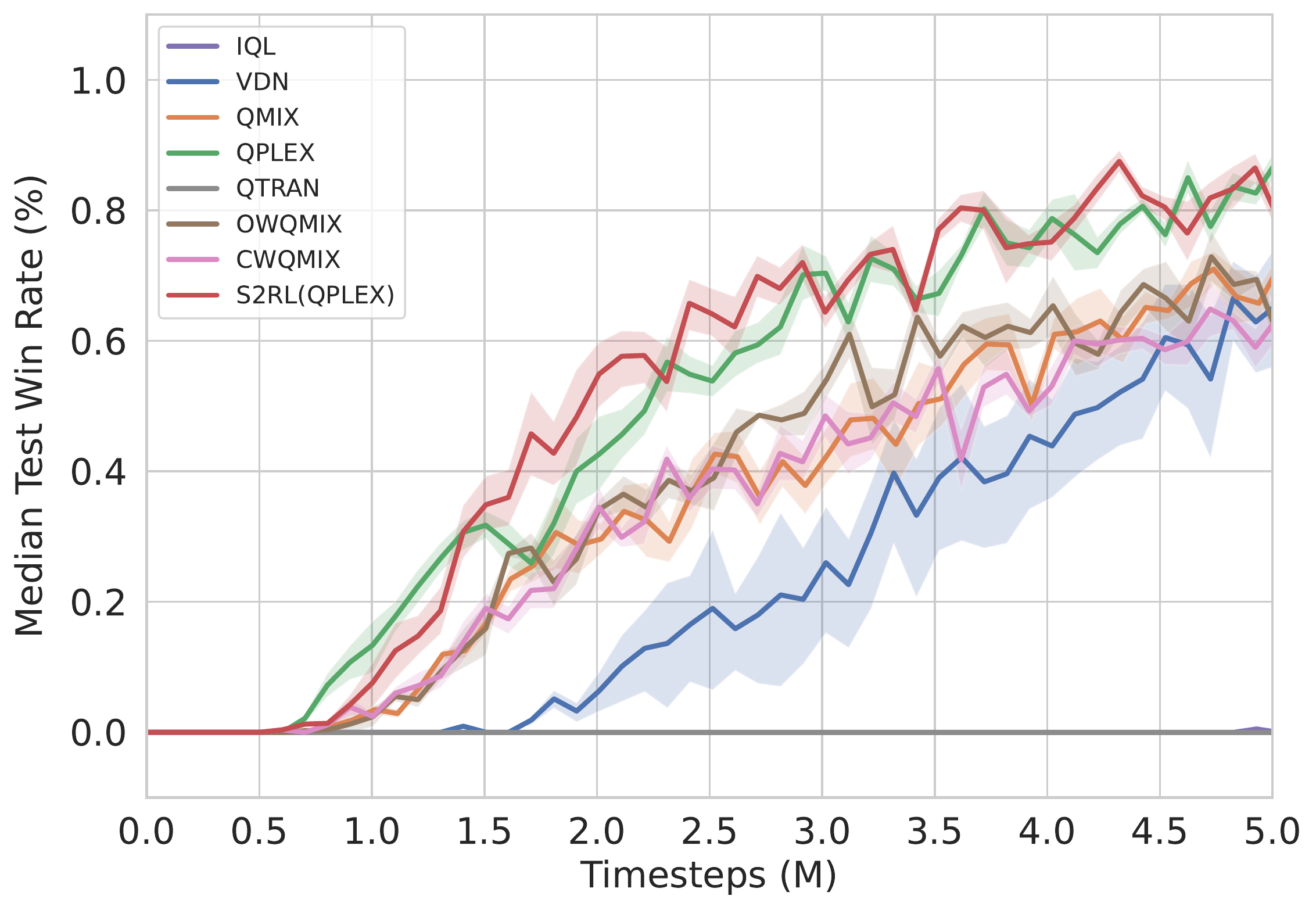}}
    
    \subfloat[corridor]{\includegraphics[scale=0.26]{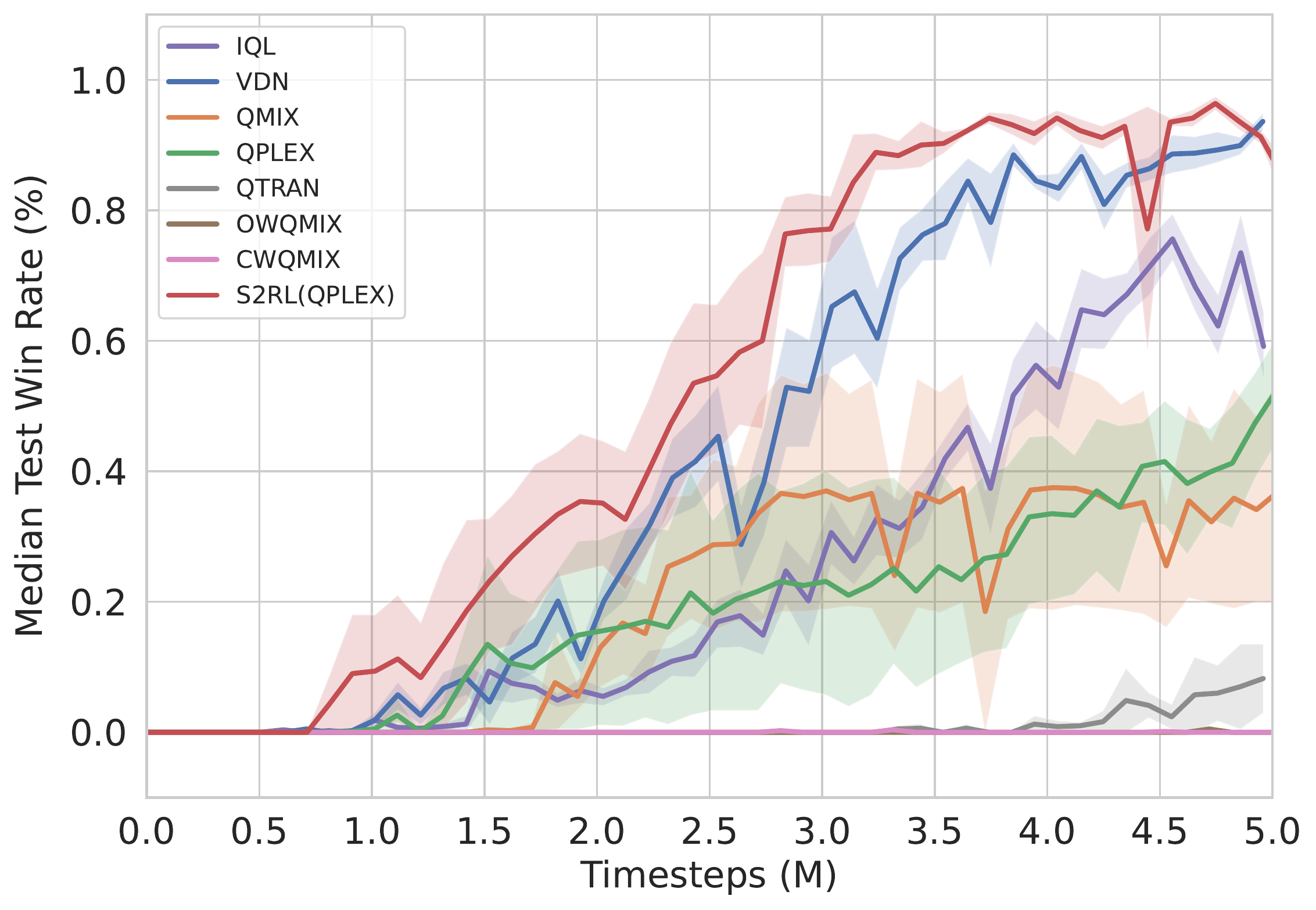}}
    \subfloat[6h\_vs\_8z]{\includegraphics[scale=0.26]{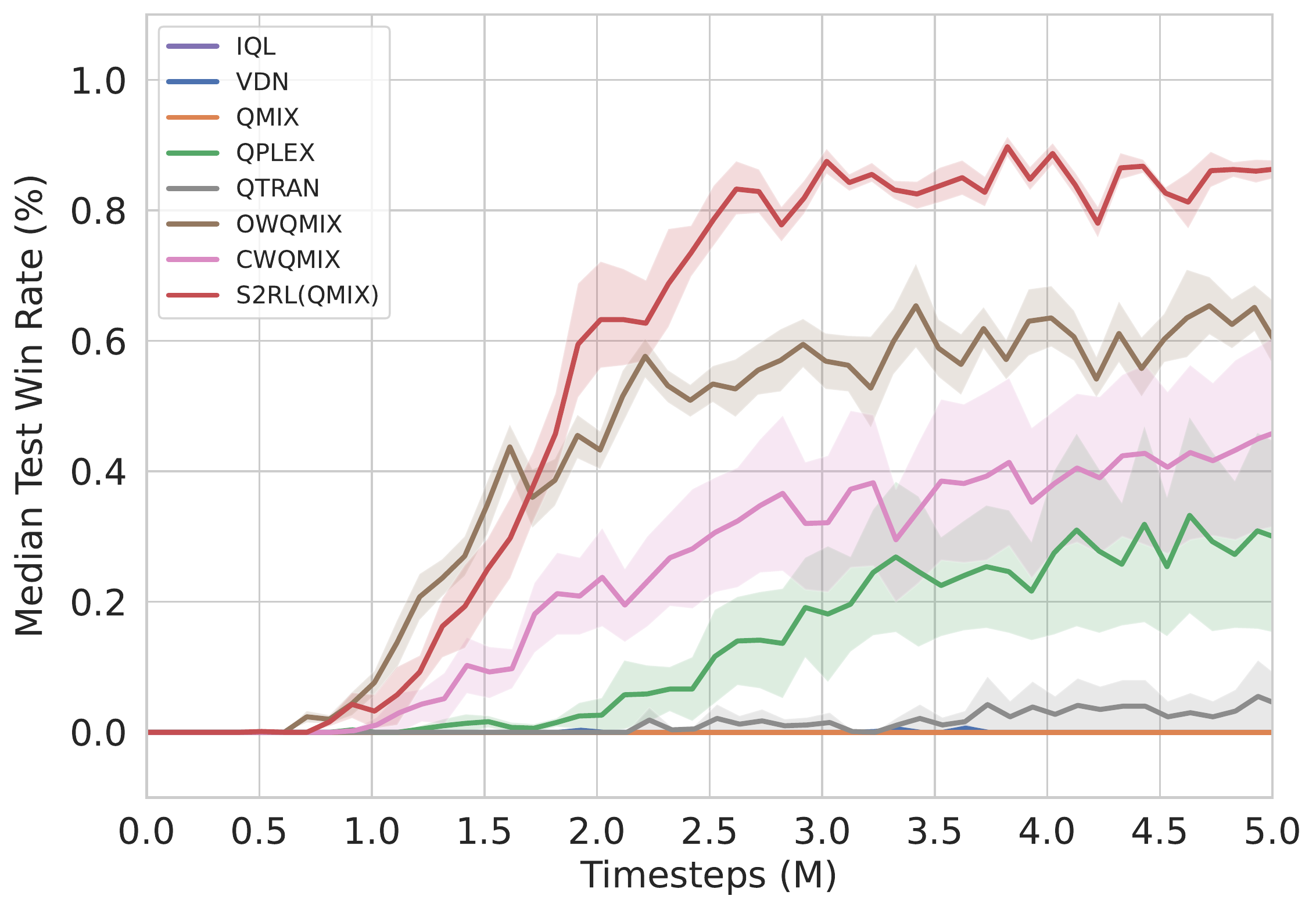}}
    \subfloat[3s5z\_vs\_3s6z]{\includegraphics[scale=0.26]{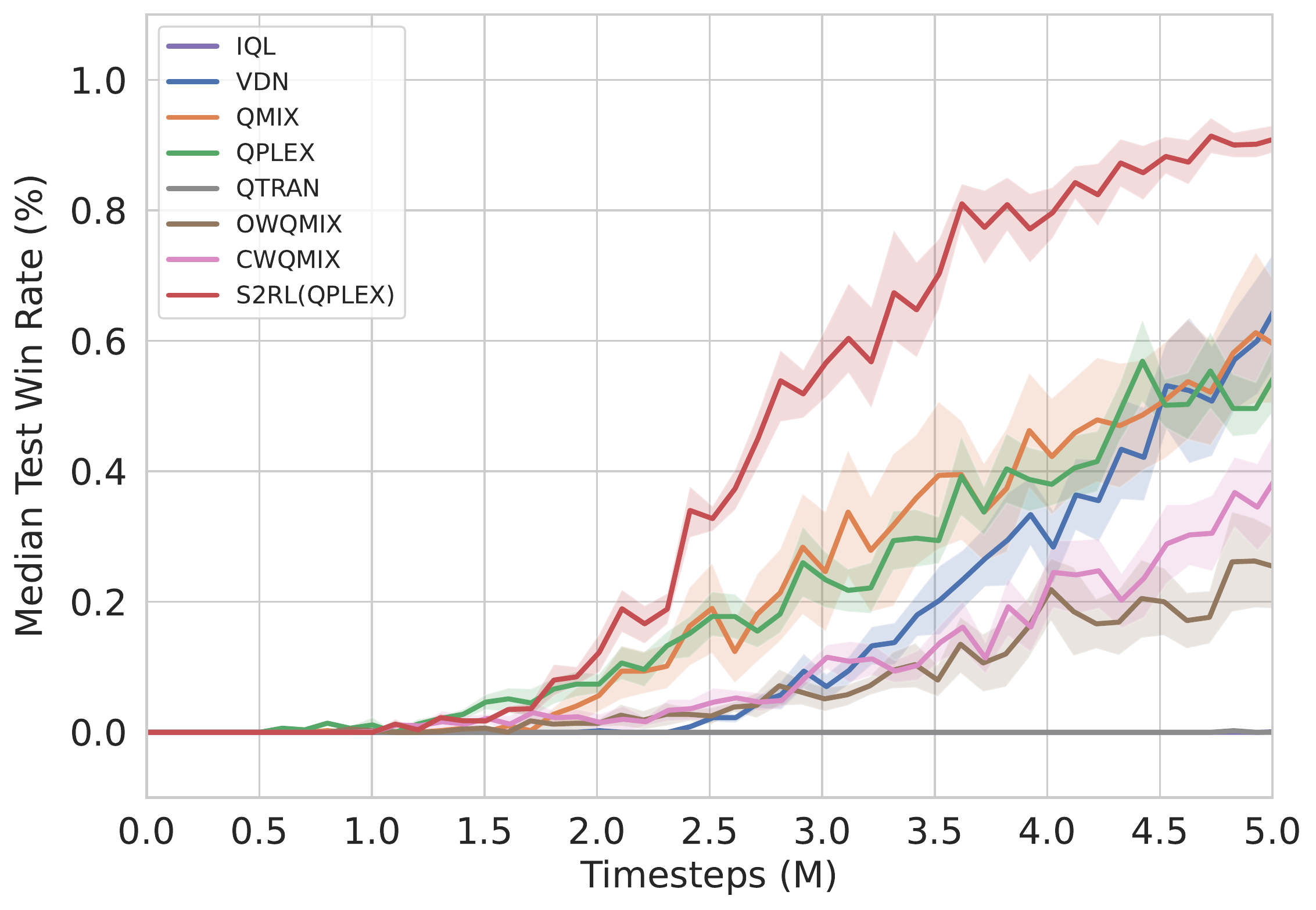}}
    \caption{Learning curves of our S2RL and baselines on one easy map (3s5z), one hard map (3s\_vs\_5z), and 4 super-hard maps ({corridor}, {5s10z}, {6h\_vs\_8z}, {3s5z\_vs\_3s6z}).
    All experimental results are illustrated with the median ($25-75\%$ percentiles) performance and across 5 runs for a fair comparison.}
    \label{fig:baseline}
\end{figure*}

% \subsection{Overall Framework}
In our framework, S2RL services as a plug-in module in the agent utility networks. The outputs of S2RL modules are directly used for subsequent network computations. Then, each agent is trained by minimizing the total loss
\begin{equation}\label{totalloss}
    \mathcal{L}\left(\theta_{\pi}, \theta_{\rho}\right) = 
     \mathcal{L}_{td}(\theta_{\pi}, \theta_{\rho}) + \lambda  \mathcal{L}_{aux}(\theta_{\pi}, \theta_{\rho}),
\end{equation}
where $\lambda$ is a regularization parameter that controls the level of attention to critical states. Obviously, a larger $\lambda$ allows our algorithm to pay more attention to some critical states, while a smaller $\lambda$ allows for a more even distribution of attention. The overall framework is trained in an end-to-end centralized manner.
The complete algorithm is summarized in Algorithm ~\ref{algo}.

% During decentralized execution, the sparse attention module is removed, which means that agents infer by dense attention solely.

\section{EXPERIMENTS}
We conduct experiments on the StarCraft Multi-Agent Challenge (SMAC)\footnote{We use the SC2.4.10 version instead of the older SC2.4.6.2.69232. Performance is not comparable between different versions.}~\cite{SMAC} to demonstrate the effectiveness of the proposed sparse state based MARL (S2RL) method. SMAC has become a standard benchmark for evaluating state-of-the-art MARL methods, which focuses on micromanagement challenges. The setup of SMAC is that each ally entity is controlled by an individual learning agent, while the enemy entities are controlled by a built-in AI. At each time step, agents can move in four cardinal directions, stop, take no-operation, or choose an enemy to attack. Thus, if there are $n_e$ enemies in the scenario, the action space for each ally unit consists of $n_e + 6$ discrete actions. Agents aim to inflict maximum damage on enemy entities to win the game. Therefore, proper tactics such as focusing fire and covering attack are required during battles. Learning these diverse interaction behaviors under partial observation is a crucial yet challenging task. In what follows, we detail the compared methods and parameter settings and then present the qualitative and quantitative performance of different methods.
% We conduct experiments  on the StarCraft Multi-Agent Challenge (SMAC)\footnote{We use SC2.4.10 version instead of the older SC2.4.6.2.69232. Performance is not comparable across versions.}~\cite{SMAC} to demonstrate the effectiveness of the proposed sparse state based MARL (S2RL) method.
% SMAC has become a common-used benchmark for evaluating state-of-the-art MARL approaches, which focuses on micromanagement challenges.
% The setup of SMAC is that each of the ally entities is controlled by an individual learning agent, and enemy entities are controlled by a built-in AI. 
% At each timestep, agents can move in four cardinal directions, stop, take noop (do nothing), or select an enemy to attack. Therefore, if there are $n_e$ enemies in the map, the action space for each ally unit consists of $n_e + 6$ discrete actions.
% Agents aims to inflict maximum damage on the enemy entities for winning the game.
% Hence proper tactics such as focusing fire and avoiding overkill are required during battles. Learning these diverse interaction behaviors under partial observation is a challenging task. In the following, we will introduce the environments and the parameters, and then both qualitative and quantitative performances of different methods.

\subsection{Comparison Methods and Training Details}
Our method is compared with several baseline methods, including IQL, VDN~\cite{VDN}, QMIX~\cite{QMIX}, QTRAN~\cite{QTRAN}, QPLEX~\cite{QPLEX}, CWQMIX and OWQMIX ~\cite{WQMIX}.
Our S2RL implementation uses VDN, QMIX and QPLEX as an integrated architecture to verify its performance, called S2RL~(VDN), S2RL~(QMIX) and S2RL~(QPLEX).
These three SOTA methods are chosen for their robust performance in different multi-agent scenarios, while S2RL can also be easily applied to other frameworks.

% Our methods are compared with several baseline methods, including IQL, VDN~\cite{VDN}, QMIX~\cite{QMIX}, QTRAN~\cite{QTRAN}, QPLEX~\cite{QPLEX}, CWQMIX and OWQMIX~\cite{WQMIX}. 
% Our S2RL implementation uses VDN, QMIX and QPLEX as incorporated architecture to validate its performance, termed as S2RL(VDN), S2RL(QMIX), S2RL(QPLEX).
% These three SOTA methods were selected for their robust performance across different multi-agent scenarios, while S2RL can be easily applicable to other frameworks.

We adopt the Python MARL framework (PyMARL)~\cite{SMAC} to run all experiments.
The hyperparameters of the baseline methods are the same as those in PyMARL to ensure comparability.  The regularization parameter in \eqref{totalloss} is set to $\lambda = 1$.
For all experiments, the optimization is conducted using RMSprop with a learning rate of $5 \times 10^{-4}$, a total timestep of $2$M, a smoothing constant of $0.99$, and no momentum or weight decay. For exploration, we use $\epsilon-$greedy with $\epsilon$ annealed linearly from $1.0$ to $0.05$ over $50K$ time steps and kept constant for the rest of the training. For four super hard exploration maps ({6h\_vs\_8z}, {3s5z\_vs\_3s6z}, {corridor}, {5s10z}), we extend the epsilon annealing time to $500K$ and the total timestep to $5M$, and three of them ({6h\_vs\_8z}, {corridor}, {5s10z}) optimized with Adam for both series of S2RL and all the baselines and ablations. Batches of $32$ episodes are sampled from the replay buffer, and all tested methods are trained end-to-end on fully unrolled episodes. All experiments on the SMAC benchmark use the default reward and observation settings of the SMAC benchmark~\cite{SMAC}. All experiments in this section were carried out with $5$ different random seeds on NVIDIA GTX V100 GPU.

% , and we randomly select $5$ experimental results for plotting the learning curves.

\begin{figure*}[h]
    \centering
    \subfloat[3s5z]{\includegraphics[scale=0.26]{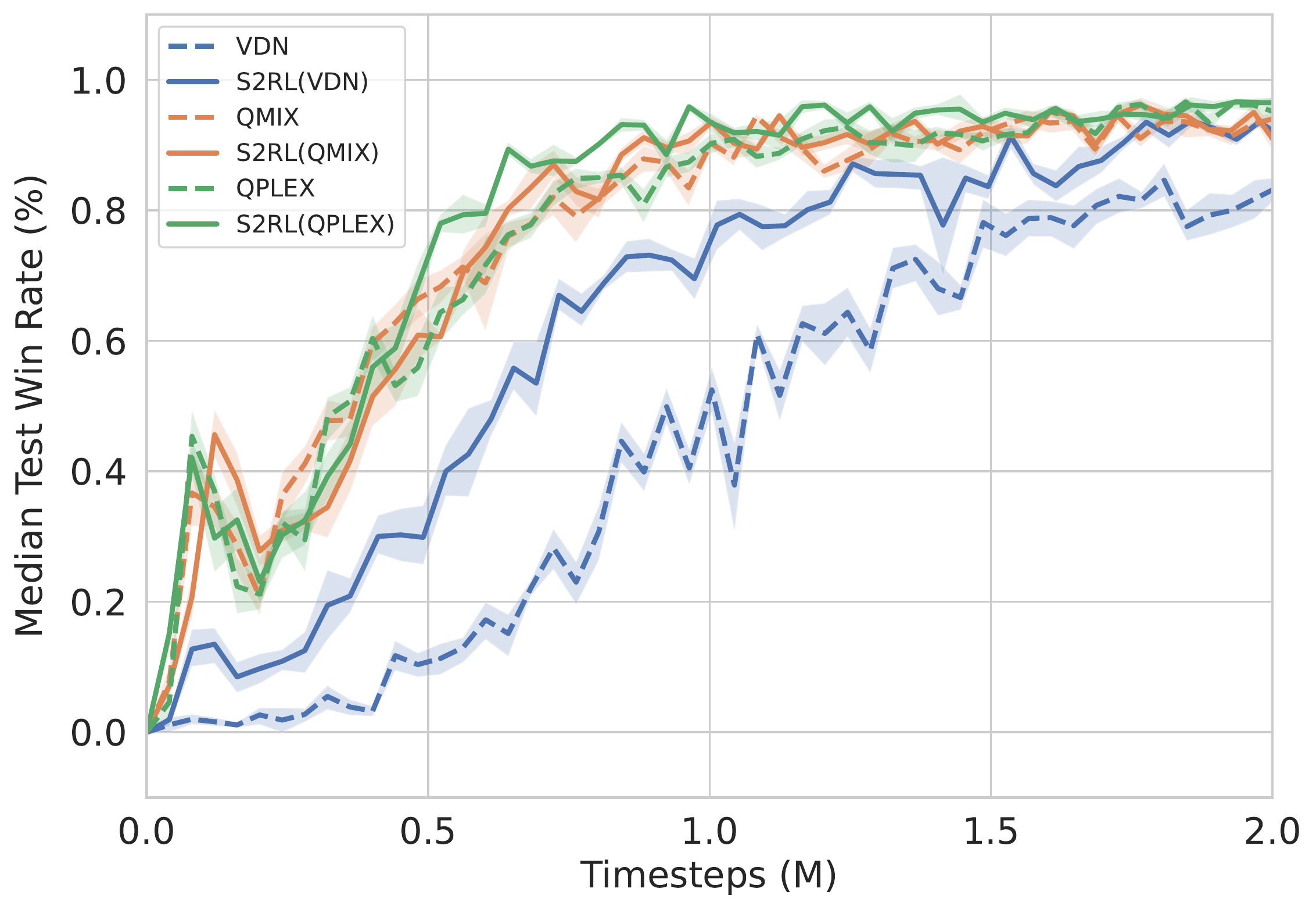}}
    \subfloat[3s\_vs\_5z]{\includegraphics[scale=0.26]{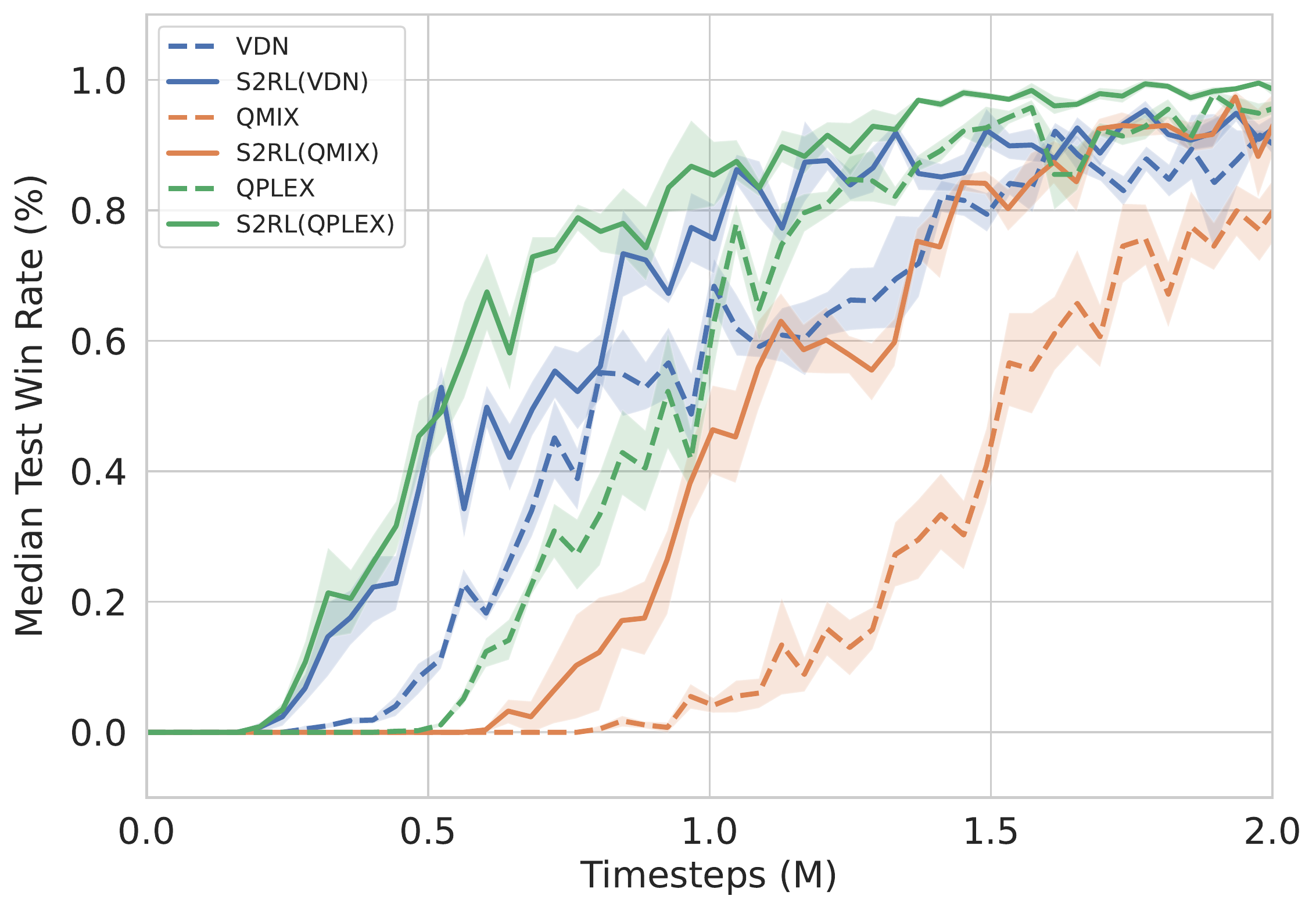}}
    \subfloat[5s10z]{\includegraphics[scale=0.26]{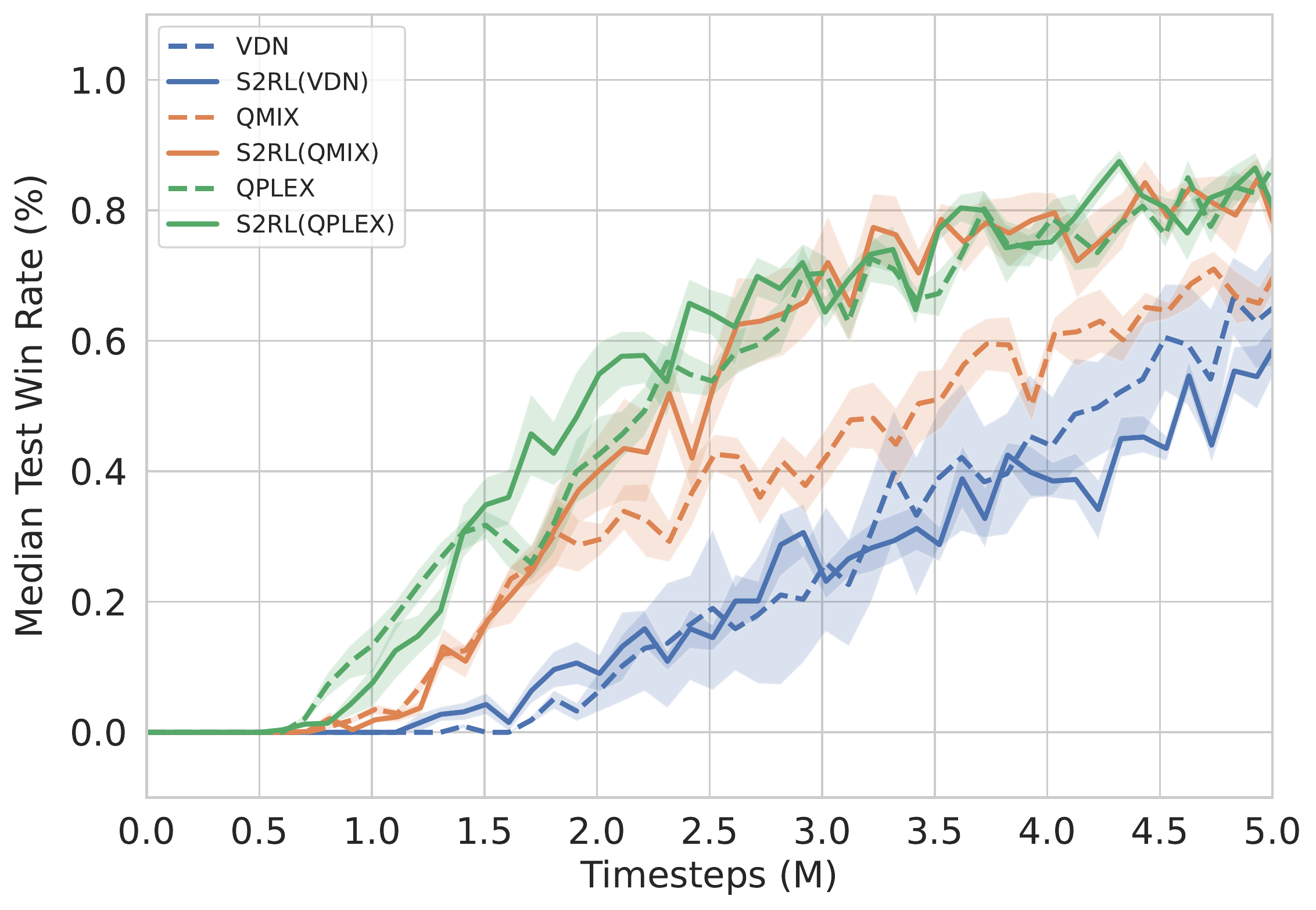}}
    
    \subfloat[corridor]{\includegraphics[scale=0.26]{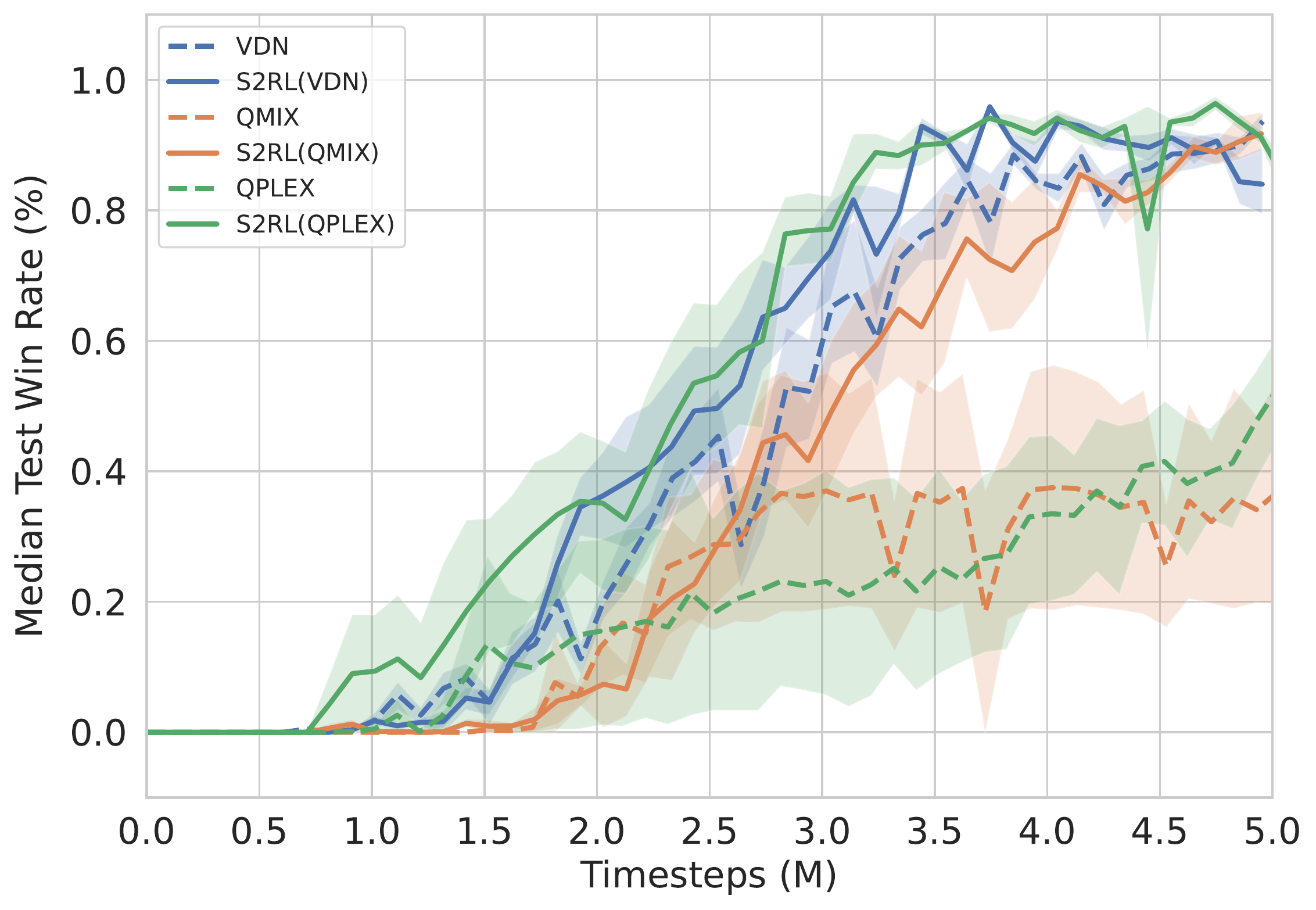}}
    \subfloat[6h\_vs\_8z]{\includegraphics[scale=0.26]{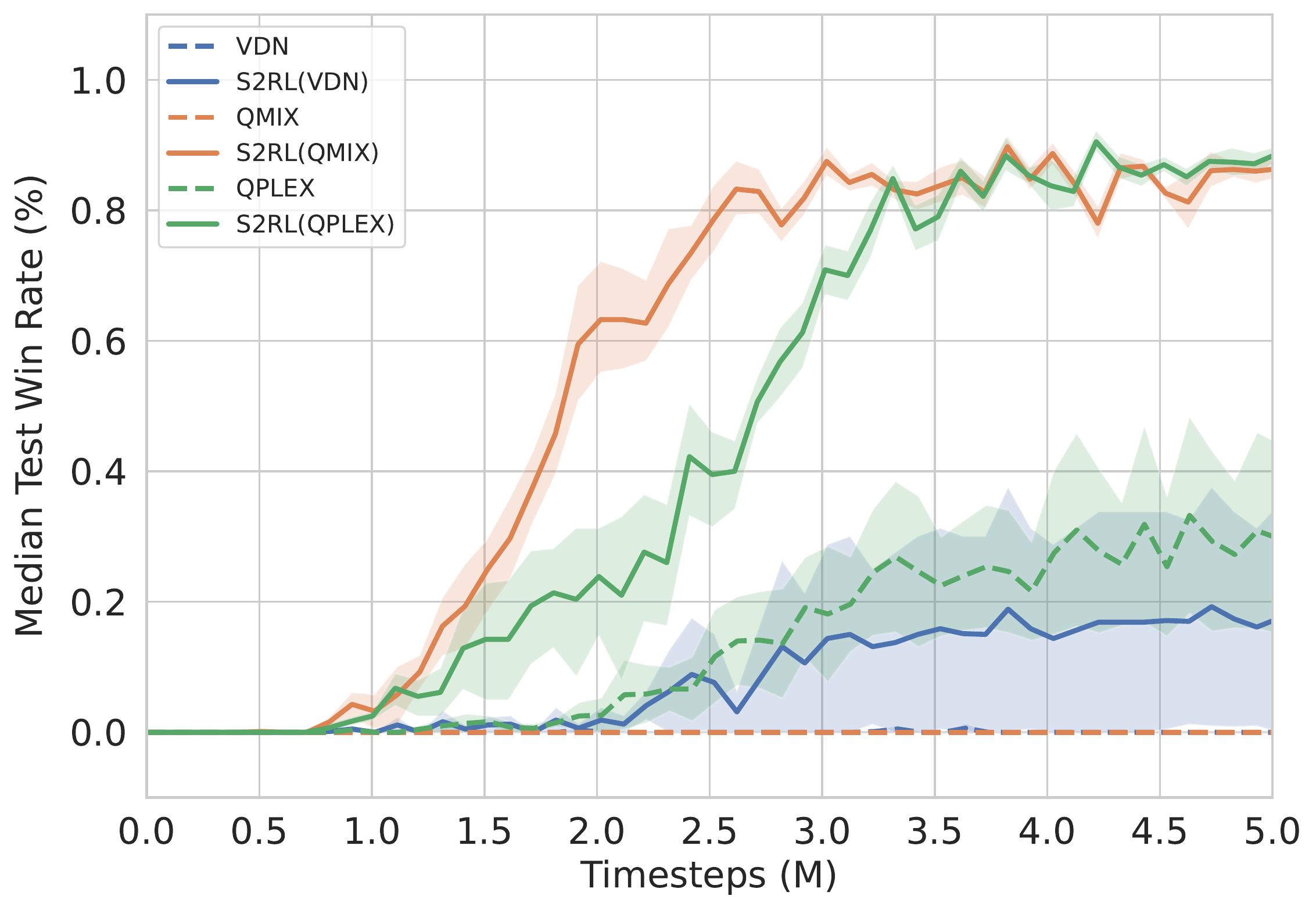}}
    \subfloat[3s5z\_vs\_3s6z]{\includegraphics[scale=0.26]{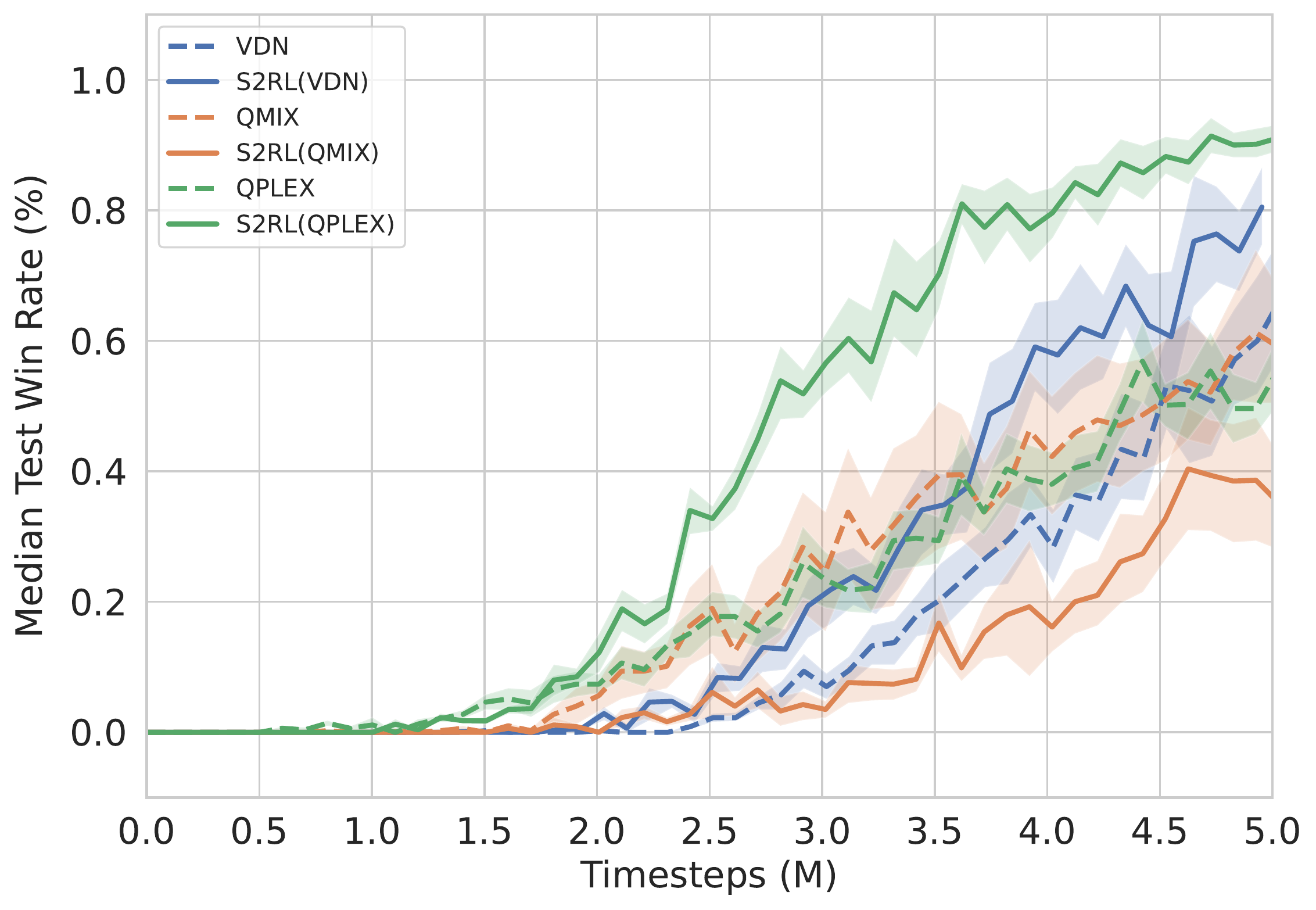}}
    \caption{The performance comparison between the vanilla methods and their S2RL variants. We integrate the proposed S2RL framework with VDN, QMIX and QPLEX.}
    \label{fig:plugin}
\end{figure*}

\begin{figure*}[h]
    \centering
    \subfloat[5s10z]{\includegraphics[scale=0.26]{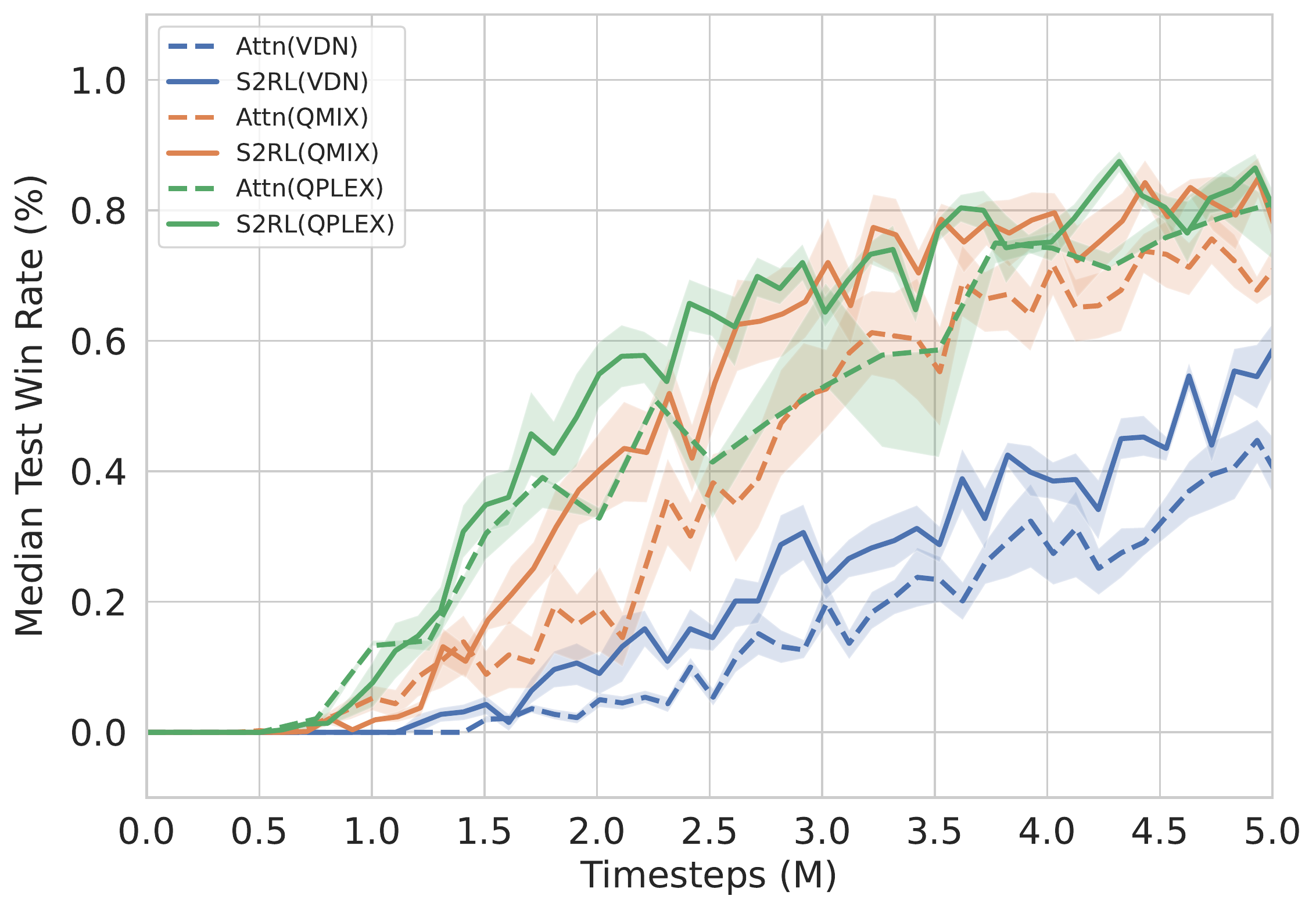}}
    \subfloat[6h\_vs\_8z]{\includegraphics[scale=0.26]{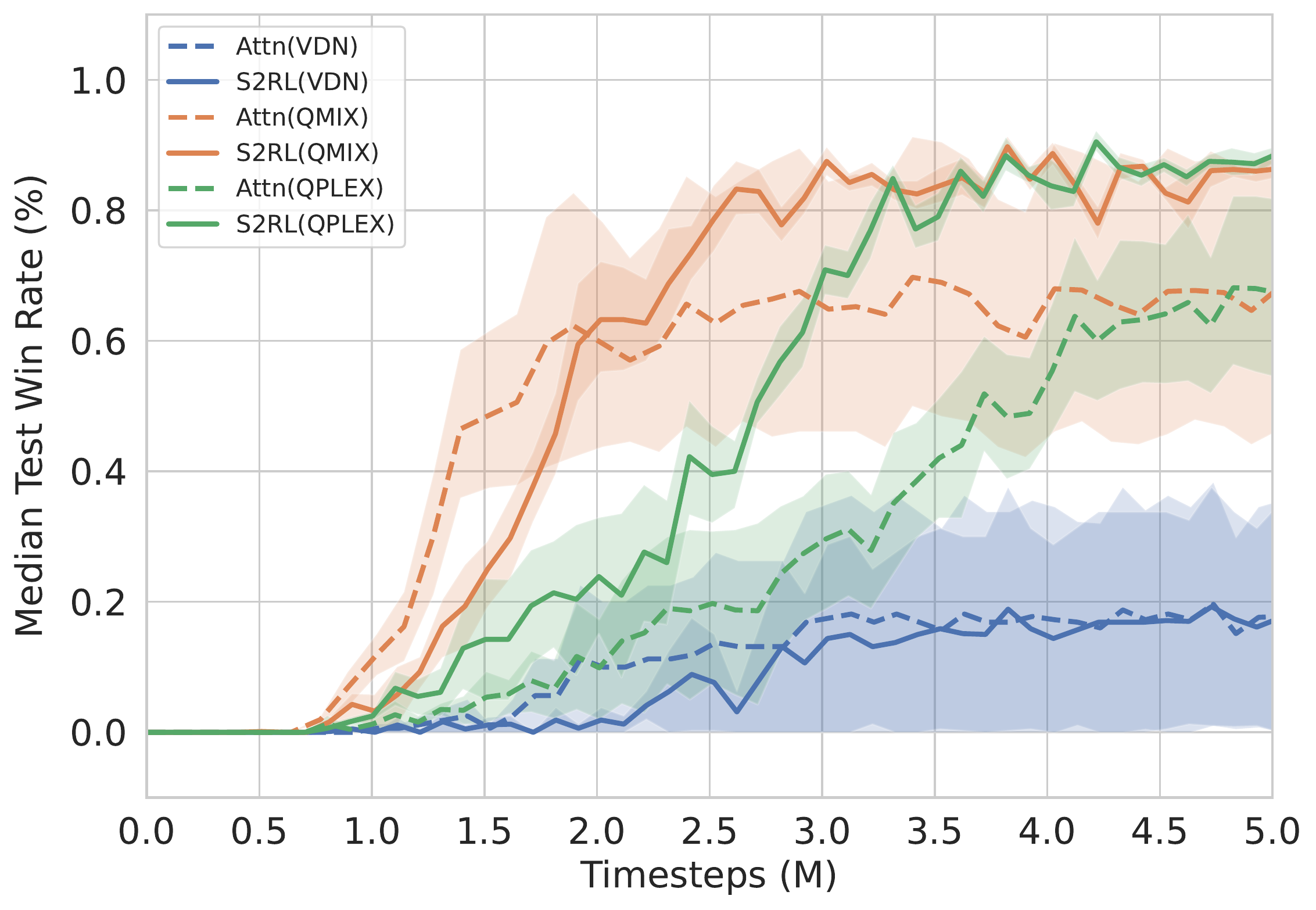}}
    \subfloat[3s5z\_vs\_3s6z]{\includegraphics[scale=0.26]{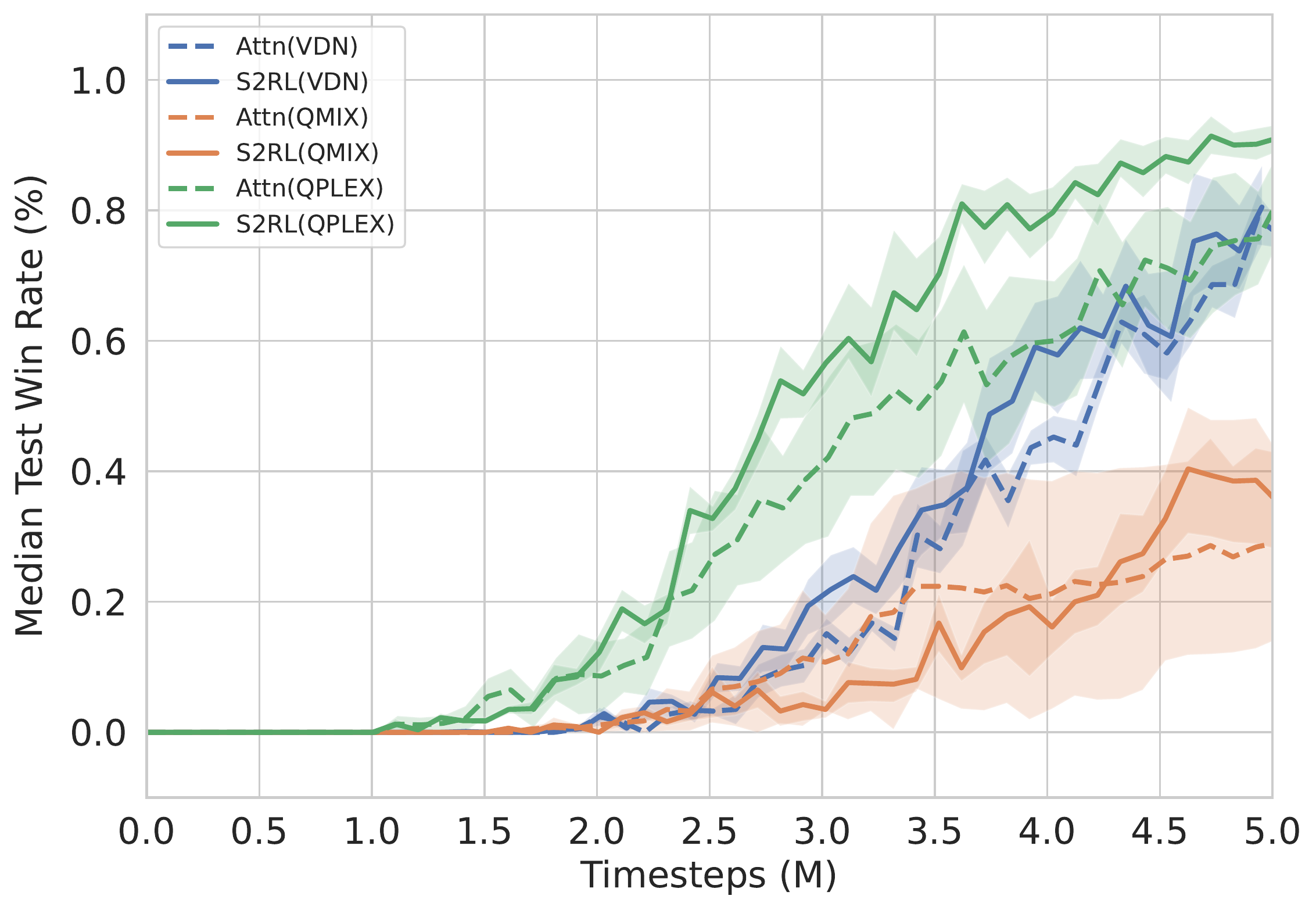}}
    \caption{Ablation studies regarding component of dense attention and auxiliary sparse attention.}
    \label{fig:ablation}
\end{figure*}

\subsection{Overall Results}
To demonstrate the efficiency of our proposed method, we conduct experiments on 6 challenging SMAC scenarios, which are classified into \textbf{Easy} (\emph{3s5z}), \textbf{Hard} (\emph{3s\_vs\_5z}) and \textbf{Super-Hard} (\emph{6h\_vs\_8z}, \emph{3s5z\_vs\_3s6z}, \emph{corridor}, \emph{5s10z}). All of these scenarios are heterogeneous, where each army is composed of more than one entity type. It is worth mentioning that MARL algorithms are harder to converge on hard and super-hard maps and therefore need to focus more on important entities to speed up convergence. In this way, we are more interested in the performance of our method on these maps.
% Since hard and super hard maps are more typically needed to learn which entities within the observation have a greater impact on the decision, which is the motivation of the S2RL, we are especially interested in the performance of our method on these maps.
% We compare the proposed S2RL(VDN), S2RL(QMIX), S2RL(QPLEX) with various MARL baselines: IQL, VDN, QMIX, QTRAN, QPLEX, OWQMIX and CWQMIX. 

Figure~\ref{fig:baseline} shows the overall performance of the tested algorithms in different scenarios. The results include the median performance and $25-75\%$ percentiles are shaded to avoid the effect of any outliers as recommended in~\cite{SMAC}. For the sake of demonstration, here we select the best plug-in method, referred to as S2RL in the following, to compare with other baseline algorithms.
First of all, we can see that S2RL performs best on up to all six tasks, which means our proposed method can efficiently enhance the performance of agents in different scenarios. In the easy map, some algorithms have achieved good performance, and our S2RL is not significantly ahead. In contrast, our S2RL significantly improves the learning efficiency and final performance compared to the baselines in some hard and super-hard scenarios.
% In the easy map, a serious of algorithms have achieved great performance, while our proposed S2RL significantly improves the learning efficiency and the final performance compared with baselines in both hard and super hard scenarios.
Specifically, in \emph{6h\_vs\_8z} and \emph{3s5z\_vs\_3s6z}, our S2RL consistently outperforms baselines by a large margin during training. 
% We hypothesize that agents do not require determining which entities are more critical on easy maps and can still find suitable strategies to defeat the enemy. The selection benefit brought by the sparse attention mechanism is not apparent. However, when the situation becomes more intricate, agents need to consider which entities are more critical to making decisions.
This is because the number of entities in easy maps is small, all entities are critical, and the selection gain brought by the sparse attention mechanism is not apparent. However, when the situation becomes more complex, and the agent needs to consider which entities are more critical to the decision, the benefits of the sparse attention mechanism are more pronounced.

% \paragraph{Robust Results}
In addition, to test the generalization of our method incorporated into various valued-based algorithms, we incorporate S2RL to VDN, QMIX and QPLEX respectively, and compare the final performance with vanilla agent utility networks in Figure~\ref{fig:plugin}. In general, most of the learning curves of S2RL~(VDN), S2RL~(QMIX) and S2RL~(QPLEX) achieve gratifying results superior to VDN, QMIX and QPLEX.
Besides, it is worth mentioning that our method pulls huge margins on tasks with more severe difficulties, demonstrating the effectiveness of S2RL. The experimental results show that in the super-hard map \emph{6h\_vs\_8z}, our proposed S2RL~(QPLEX) improves the win rate by almost $55\%$ compared to the naive QPLEX. Even more encouraging is that S2RL~(QMIX) can reach a win rate of $80\%$ while QMIX basically does not learn any strategy.

% In addition, the promotion of incorporating S2RL to QMIX and QPLEX is higher than VDN, which is relevant to the mixing network. We hypothesize that sparse attention mechanism select the critical entities and then discriminate their contributions to the estimation of value functions, which may promote the ability of credit assignment. Unlike QMIX and QPLEX, VDN represents joint action-value as a summation of agents’ individual Q-functions resulting in this poor representation of mixing network challenging to leverage the strengths of our approach.  

Furthermore, the promotion of incorporating S2RL into QMIX and QPLEX is higher than VDN, which reveals the importance of the mixing network. We hypothesize that the sparse attention mechanism enables the model to select critical entities and further clarify their contributions, which may promote the power of credit assignment. Unlike QMIX and QPLEX, VDN represents the joint action-value as a summation of individual Q-functions, resulting in this poor representation of the mixing network challenging to leverage the strengths of our approach.

\subsection{Ablation Study}
% \paragraph{Ablation Results}
% To demonstrate the advantage of sparse auxiliary loss to the training process of agents, we carry out ablation studies to test its contribution.
% three super hard maps (\emph{6h\_vs\_8z}, \emph{3s5z\_vs\_3s6z}, \emph{corridor})
To evaluate the advantage of sparse auxiliary loss on the agent training process, we conduct ablation studies on three super hard maps (\emph{5s10z}, \emph{6h\_vs\_8z}, \emph{3s5z\_vs\_3s6z}) to test its contribution.
Our S2RL mainly consists of two parts: (A) dense attention, denoted Attn; (B) sparse attention as an auxiliary, noted as S2RL. We apply these two components to VDN, QMIX and QPLEX utility networks and compare their performance in Figure~\ref{fig:ablation}.
% Our proposed method S2RL consists of two mainly components: (A) Dense attention, noted as Attn; (B) Sparse attention as auxiliary, noted as S2RL. We apply these two components into VDN, QMIX, QPLEX utility network and compare their performance in Figure~\ref{fig:ablation}. 
The solid curves indicate that the agents use the dense attention module to calculate the importance of different entities. The dashed curves indicate that the agents learn to use the sparse attention module as an auxiliary to teach the dense attention module.
% The solid curves represent that the agents use dense attention mudule to calculate the importance of different entities. The dotted curves represent that the agents learn to use sparse attention module as auxiliary to teach the dense attention mudule.

% Generally speaking, the advantages of using S2RL become apparent gradually in the middle and late stages of training. We hypothesize that agents can not distinguish which entity is more important when the training starts. As the training progressed, agents explore more unknown states and are gradually able to distinguish which entities are more critical. 
Generally speaking, the advantages of using S2RL gradually emerge in the middle and late stages of training. We assume that agents cannot distinguish which entity is more important at the beginning of training. As training progresses, agents explore more unknown states and are gradually able to distinguish which entities are more critical. Finally, the overall performance of agents is improved when they discard irrelevant entities.
Furthermore, we find that using sparse attention achieves more significant improvements on \emph{6h\_vs\_8z} and \emph{corridor}.
On the \emph{6h\_vs\_8z} scenario, $6$ Hydralisks face $24$ enemy Zealots, while on the \emph{corridor} scenario, $6$ Zealots face $24$ enemy Zerglings.
The controllable agents in these scenarios are homogeneous, making it easier for them to explore cooperative strategies. Moreover, using the sparse attention module helps simplify the exploration space, making S2RL more advantageous in these scenarios.

% Finally, the overall performance of agents get an improvement when they discard irrelevant entities.
% Furthermore, we find that compared with \emph{3s5z\_vs\_3s6z}, the use of sparse attention achieve more improvement on \emph{6h\_vs\_8z} and \emph{corridor}. 
% In \emph{6h\_vs\_8z} map, $6$ Hydralisks face $24$ enemy Zealots, and in \emph{corridor}, $6$ Zealots face $24$ enemy Zerglings.
% The controlable agents in these scenarios are homogeneous, which makes them easier to explore collaborative strategies. Thus, the use of sparse attention module helps simplify the search strategy space, resulting in S2RL being more advantageous in these scenarios.

% The results suggest that utilizing sparse attention mechanism as auxiliary loss can explore the diverse unknown entities of observations, which helps the agents to construct a more useful policy and achieves non-trivial performance.

\begin{figure*}[!t]
    \centering
    \subfloat[Strategy: Zealots $0$ leave the team separately to attract the attention of most enemies.]{\includegraphics[scale=0.355]{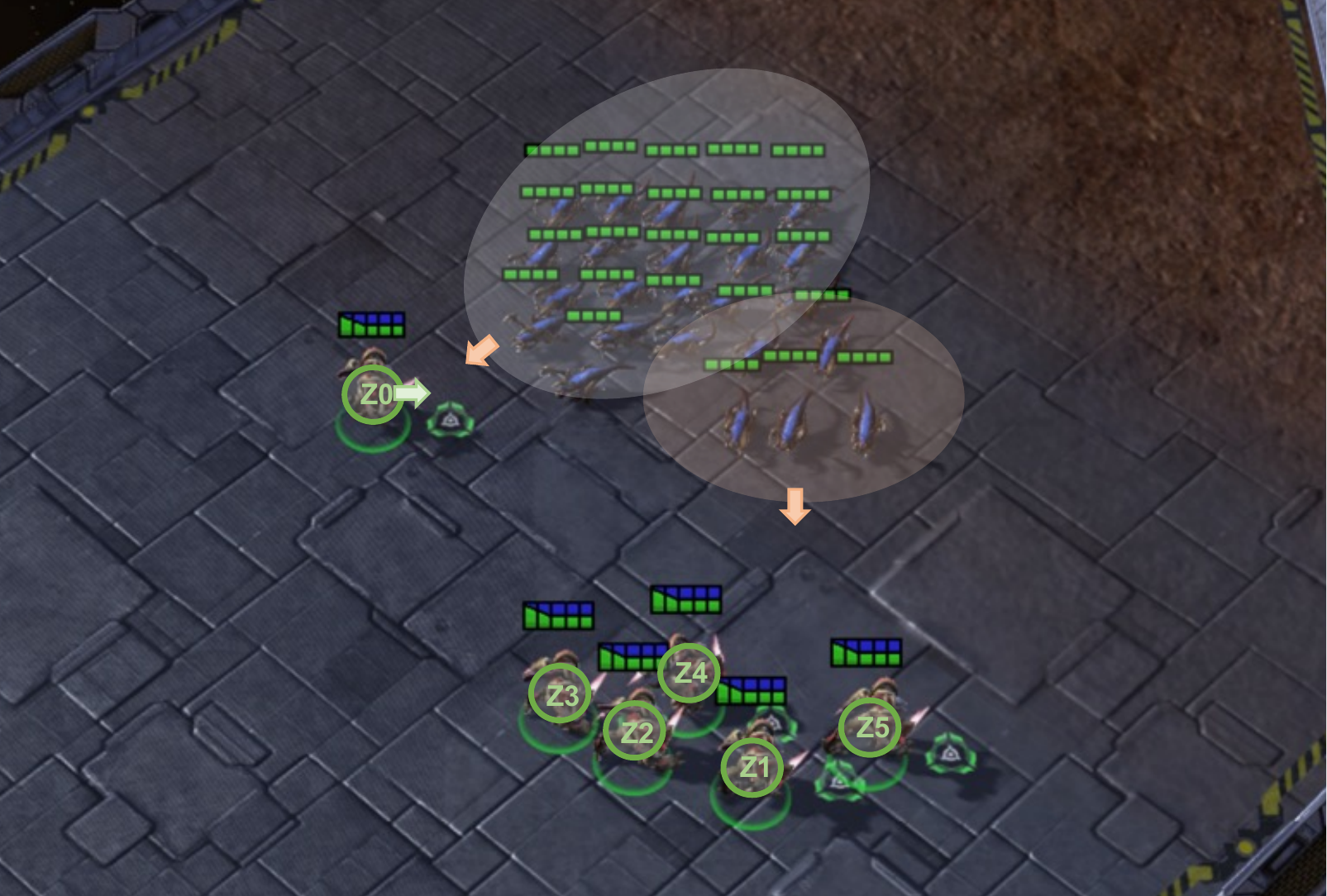}
    \label{fig:visual-a}}\;
    \subfloat[Strategy: Zealots $1$, $2$, $5$ focus fire cooperatively and Zealots $4$ attack the distant enemy to rescue his teammates. ]{\includegraphics[scale=0.355]{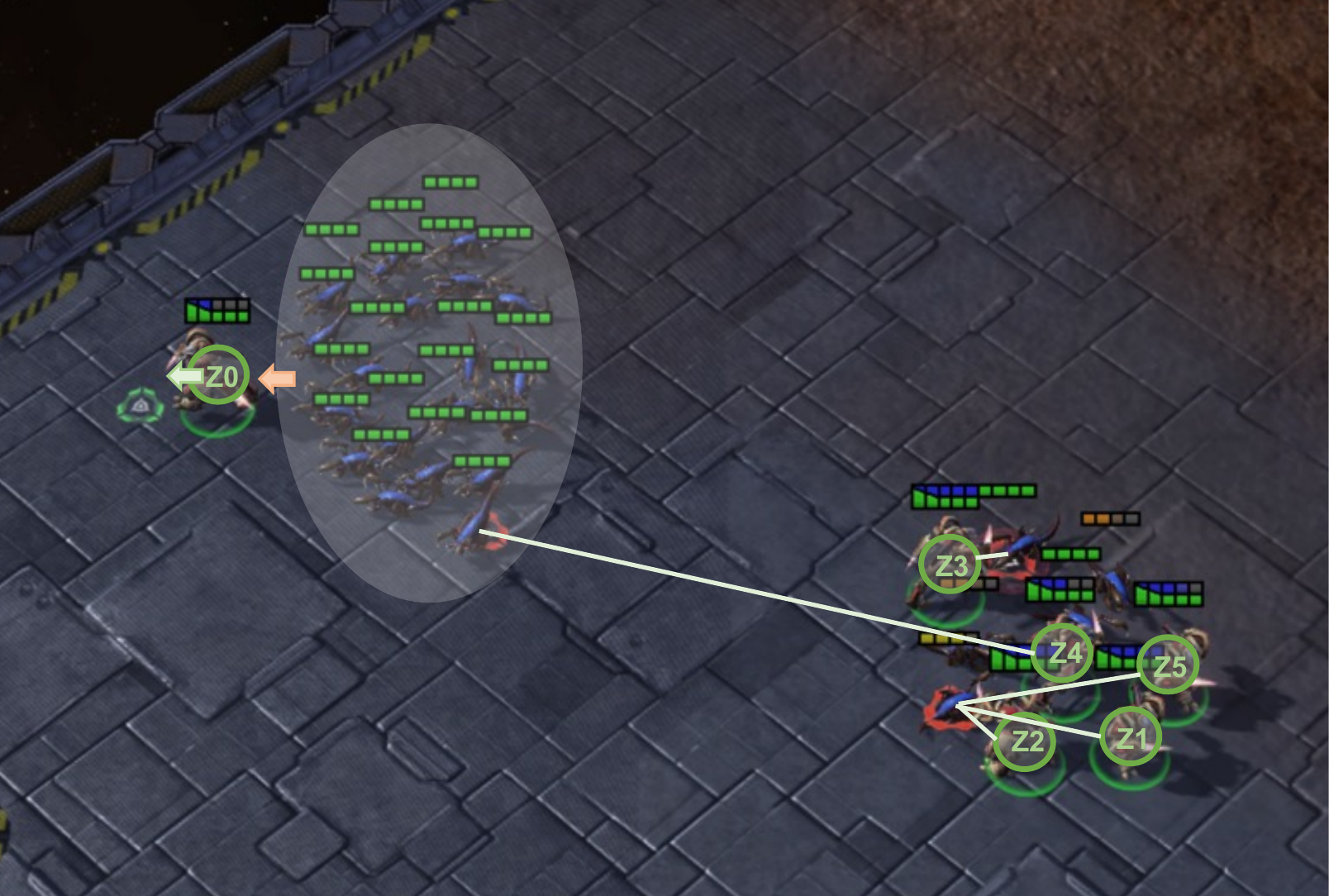}\label{fig:visual-b}}\;
    \subfloat[Strategy: Zealots $0$ keep moving to avoid being attacked and others eliminate the scattered enemies. ]{\includegraphics[scale=0.355]{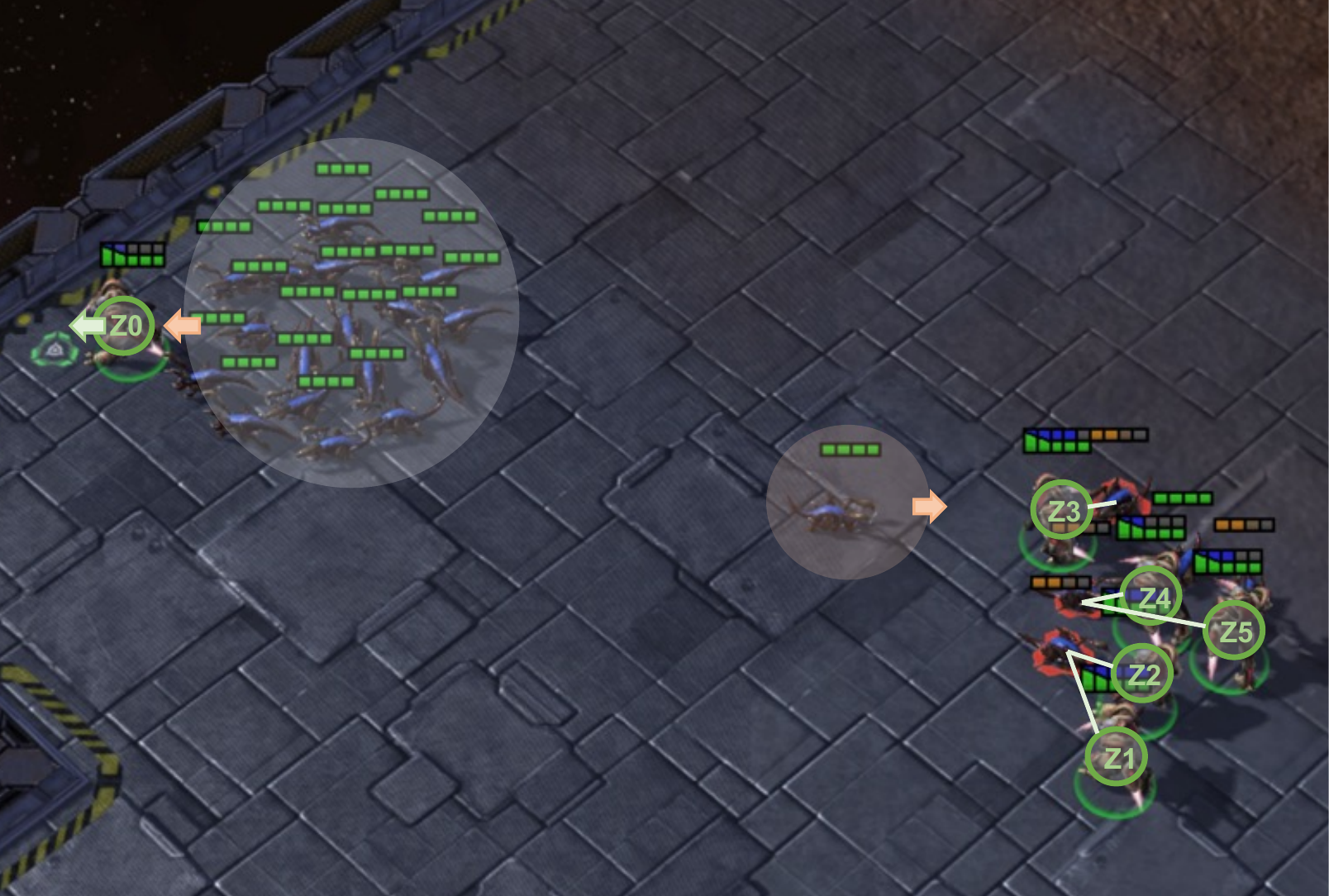}\label{fig:visual-c}}
    \caption{A visualization example of the sophisticated strategies adopted by S2RL~(QMIX) in the SMAC corridor scenario.
    In this super-hard map, ally units are 6 Zealots labeled by green circle, while enemy units are 24 Zerglings.
    Green and red shadows mark enemies attracted by ally units. Green arrows and red arrows indicate the direction in which ally units and enemy units will move, respectively. Yellow lines indicate enemies that ally units are attacking.}
    \label{fig:visual}
\end{figure*}

\subsection{Action Representations}
% Figure~\ref{fig:visual} visualize the final trained model S2RL(QMIX) on SMAC corridor map to better explain why our approach performs well. In this super hard map, six friendly Zealots are facing 24 enemy Zerglings. The disparity in quantity means our agents need to learn cooperative strategies, such as moving, pulling, focusing fire, etc. Otherwise, agents are doomed to lose if they gather together. 

Figure ~\ref{fig:visual} visualizes the final trained model S2RL~(QMIX) on the SMAC corridor scenario to better explain why our method performs well. In this super-hard scenario, 6 friendly Zealots face 24 enemy Zerglings. 
The disparity in quantity means that our agents need to learn cooperative strategies such as moving, pulling and focusing fire. Otherwise, agents are doomed to lose if they gather together. 

% As shown in Figure~\ref{fig:visual}(a), at the beginning of the game, Zealots $0$, whose route is highlighted in green, becomes a warrior leaving the team separately to attract the attention of most enemies in the green oval. Thus the others Zealots can eliminate a small part of the enemies in the red oval with higher probability of winning.

As shown in Figure~\ref{fig:visual}(a), the game starts with the Zealots $0$ highlighted in green as a warrior, leaving the team separately to grab the attention of most of the enemies in the green oval. Thus other zealots can eliminate a small number of enemies in the red oval with a high probability of winning. In Figure~\ref{fig:visual}(b), we can see that Zealots $1$, Zealots $2$ and Zealots $5$ are focusing fire on the enemy, thus speeding up the eradication of the enemy. In the meanwhile, Zealots $4$ stands out to attack enemies surrounding their teammates from a distance. These sophisticated strategies reflect that Zealots $4$ has a better sense of the situation and knows what it should do to protect its teammates. In the next time step, we recognize that Zealots $0$ is constantly moving to avoid being attacked, and the enemy marked by the red oval is successfully drawn and walking towards our team (see Figure~\ref{fig:visual}(c)). Although doomed to sacrifice, Zealots $0$ gives teammates plenty of time to annihilate scattered enemies and rescue Zealots $0$. All in all, S2RL can effectively allow agents to immediately focus on critical entities and make decisions, especially in more intricate scenarios.

% In Figure~\ref{fig:visual-b} we can noted that Zealots $1$, Zealots $2$ and Zealots $5$ are focusing fire to attack the enemy, which speeds up the eradication of the enemy. In the meanwhile, Zealots $4$ stands out to attack the enemy who surrounded his teammates in the distance. These sophisticated strategies reflect that Zealots $4$ has a better judgment of the situation and knows what he should do to protect his teammates.

% In the next time step, we recognize that Zealots $0$ moved constantly to avoid being attacked and the enemy labeled in the red oval is being attracted successfully and coming towards our team, as shown in Figure~\ref{fig:visual-c}. Thus, we have enough time to Although doomed to sacrifice, he brings enough time for his teammates to annihilate the scattered enemies and rescue Zealots $0$.  To sum up, S2RL is effective for agents to focus immediate attention on critical entities and make decisions, especially in the more intricate scenarios.

\section{Related works}
\subsection{Value-based Methods in MARL}

Recently, value-based methods have been applied to multi-agent scenarios to solve complex Markov games and have achieved significant algorithmic progress. VDN~\cite{VDN} represents the joint action-value as a summation of individual value functions. Due to its poor expression factorization, QMIX~\cite{QMIX} improves VDN~\cite{VDN} by using a mixing network for nonlinear aggregation while maintaining the monotonic relationship between centralized and individual value functions. Moreover, weighted QMIX~\cite{WQMIX} adapts a twin network and encourages underestimated actions to alleviate the risk of suboptimal outcomes. The monotonic constraints of QMIX and similar methods lead to provably poor exploratory and suboptimal properties. To address the structural limitations, QTRAN~\cite{QTRAN} constructs regularizations with linear constraints and relaxes them with a $\ell_2$-norm penalty to improve tractability, but its constraints are computationally intractable. MAVEN~\cite{MAVEN} relaxes QTRAN~\cite{QTRAN} by two penalties and introduces a hierarchical model to coordinate diverse explorations among agents. In \cite{QPLEX}, a duplex dueling network architecture is introduced for factoring joint value functions, which achieves state-of-the-art on a range of cooperative tasks. Additionally, some more advanced methods~\cite{ROMA,RODE} introduce role-oriented frameworks to decompose complex MARL tasks. In general, these methods mainly focus on aggregating local agent utility networks into a central critic network, while our method improves the structure of individual agent networks for more robust performance.

\subsection{Attention Mechanism in MARL}
Recently, attention models are increasingly adopted in MARL algorithms~\cite{attention,SENet,GAT}, since the attention mechanism is effective in extracting communication channels, representing relations, and incorporating information in large contexts. 
ATOC~\cite{ATOC} and MAAC~\cite{MAAC} process messages from other agents differently through the attention layer according to their state-dependent importance. 
SparseMAAC~\cite{sparsemaac} extends MAAC~\cite{MAAC} with sparsity by directly replacing the softmax activation function in the attention mechanism with $\gamma$-sparsemax.
% It calculates the importance of different agents observations and actions, while we mainly focus on selecting key entities of individual observations.}
In addition, TarMAC \cite{Tarmac} utilizes a sender-receiver soft attention mechanism and multiple rounds of cooperative reasoning to allow targeted continuous communication between agents. Then CollaQ\cite{zhang2020multi} considers the use of attention mechanisms to handle a variable number of agents to solve the problem of dynamic reward distribution. 
Qatten~\cite{Qatten} employs an attention mechanism to compute the weights of local action-value functions and mix them to approximate the global Q-value. EPC~\cite{EPC} utilizes an attention mechanism to combine embeddings from different observation-action encoders. REFIL~\cite{REFIL} uses attention in QMIX to generate a random mask group of agents. UPDET~\cite{UPDeT} decouple the policy distribution from intertwined input observations with the help of a transformer mechanism. 
Moreover, G2ANet~\cite{G2ANet} and HAMA~\cite{HAMA} construct the relationship between agents as a graph and utilize attention mechanisms to learn the relationship between agents. 
% However, these existing attention mechanisms compute the importance weights of all entities, ignoring the dynamic importance among them. In this case, all participants are assigned scores according to a dense fully connected graph, which forces agents to perceive all entities.
However, most of these existing attention mechanisms compute the importance weights of all entities. In this case, all participants are assigned scores according to a dense fully connected graph, which forces agents to perceive all entities. 
SparseMAAC takes sparsity into account,  but it ignores that directly applying the sparse attention mechanism will disrupt sufficient exploration and push the algorithm towards suboptimal policies.
In this paper, agents learn to perceive more critical entities of observation in the decision-making process while all observation information is preserved.

\section{CONCLUSION}
% In this work, we investigate how cooperative MARL agents can benefit from extracting significant entities in the observation. We design a novel Sparse State based MARL (S2RL) algorithm which utilizes sparse attention mechanism as an auxiliary way to select the critical entities and neglect the extraneous information. Moreover, S2RL can be readily incorporated into various action-valued-based architectures such as VDN, QMIX, QPLEX, etc. Experimental results on StarCraft II micromanagement benchmarks and across different value-based backbones demonstrate that our method significantly outperforms existing cooperative MARL algorithms and achieves state-of-the-art. It is worth mentioning that our method pulls huge margins on tasks with more severe difficulties, demonstrating the effectiveness of S2RL.
In this work, we investigate how cooperating MARL agents benefit from extracting significant entities from observations. We design a novel sparse state based MARL algorithm that utilizes a sparse attention mechanism as an auxiliary way to select critical entities and ignore extraneous information. Moreover, S2RL can be easily integrated into various value-based architectures such as VDN, QMIX, QPLEX, etc. Experimental results on the StarCraft II micromanagement benchmark and different value-based backbones demonstrate that our method significantly outperforms existing collaborative MARL algorithms and achieves state-of-the-art. It is worth mentioning that our method pulls huge margins on complex tasks, demonstrating the effectiveness of S2RL.
It could be interesting to investigate the grouping between cooperating agents through sparseness for future work.

% It could be interesting to investigate grouping among the cooperative agents by sparsification for future work. With this inferred knowledge, agents with the same sparse attention distribution can be regarded as a group that may conduct similar behavior patterns. In this way, agents can learn from both individual and population perspectives.
% It could be interesting to investigate the grouping between cooperating agents through sparsification for future works. Armed with this inferred knowledge, agents with the same sparse attention distribution can be regarded as a group likely to engage in similar behavior patterns. In this way, agents can learn from the perspective of individuals and groups.

\begin{acks}
This work was supported by the National Key Research and Development Project of China (2021ZD0110400 $\&$ 2018AAA0101900), National Natural Science Foundation of China (U19B2042), The University Synergy Innovation Program of Anhui Province (GXXT-2021-004), Zhejiang Lab (2021KE0AC02), Academy Of Social Governance Zhejiang University, Fundamental Research Funds for the Central Universities (226-2022-00064 $\&$ 226-2022-00142), Artificial Intelligence Research Foundation of Baidu Inc., Program of ZJU and Tongdun Joint Research Lab, Shanghai AI Laboratory (P22KS00111).
\end{acks}

\balance
\bibliographystyle{ACM-Reference-Format}
\bibliography{sample-base}

% \appendix

% \section{appendix}

% \subsection{appendix}

\end{document}